\documentclass[11pt]{article}

\usepackage[preprint]{acl}

\usepackage{times}
\usepackage{latexsym}
\usepackage{calc}

\usepackage[T1]{fontenc}
\usepackage[most]{tcolorbox}
\usepackage{geometry}
\usepackage[utf8]{inputenc}
\usepackage{hyperref}
\usepackage{microtype}

\usepackage{inconsolata}

\usepackage{graphicx}
\usepackage{float}
\usepackage{enumitem}
\usepackage{amsmath}
\usepackage{amssymb}
\usepackage{mathtools}
\usepackage{amsthm}
\usepackage{booktabs}
\usepackage{multirow}
\usepackage{subcaption}

\theoremstyle{plain}

\theoremstyle{definition}

\theoremstyle{remark}

\newtcolorbox{evaluationbox}[1]{
    enhanced,
    width=\textwidth, 
    colback=gray!20,           % 主体背景颜色（浅灰色）
    colframe=black,           % 边框颜色
    colbacktitle=black,       % 标题栏背景颜色
    coltitle=white,           % 标题文字颜色
    fonttitle=\ttfamily\bfseries\large, % 标题字体设置
    fontupper=\ttfamily,      % 正文字体设置（等宽字体）
    arc=5pt,                  % 圆角半径
    outer arc=5pt,
    left=15pt,
    right=15pt,
    top=10pt,
    bottom=10pt,
    boxrule=1pt,              % 边框粗细
    title=#1,                 % 标题内容
    drop shadow=black!50!white % 可选：添加阴影（如不需要可删除）
}

\newtcolorbox{mathanalysisbox}{
    enhanced,
    width=\textwidth,         % 横跨全页面宽度
    colback=orange!5!white,   % 模拟图片中的米色/淡橙色背景
    colframe=black!80,        % 深色边框
    fontupper=\ttfamily\small, % 使用等宽字体
    arc=6pt,                  
    outer arc=6pt,
    left=15pt,
    right=15pt,
    top=15pt,
    bottom=15pt,
    boxrule=1pt,
    boxsep=2pt,
    drop shadow=black!20!white
}

\title{Learning to Reason with Insight for Informal Theorem Proving}

\author{Yunhe Li${}^{*1}$ \quad Hao Shi${}^{*2}$ \quad Bowen Deng${}^{*1}$ \quad  Wei Wang${}^{*3}$ \quad Mengzhe Ruan${}^1$ \quad Hanxu Hou${}^4$ \\
\textbf{Zhongxiang Dai${}^5$ \quad Siyang Gao${}^1$ \quad Chao Wang${}^{13}$ \quad Shuang Qiu${}^{\dag 1}$ \quad Linqi Song${}^{\dag 1}$} \\
  ${}^1$City University of Hong Kong\quad ${}^2$Tsinghua University \quad ${}^3$Ke Holdings Inc.\\
    ${}^4$Shenzhen University of Advanced Technology\quad ${}^5$Chinese University of Hong Kong, Shenzhen\\
  \texttt{\{uuen.li,bowdeng2-c,cs.mzr\}@my.cityu.edu.hk} \quad \texttt{shih22@mails.tsinghua.edu.cn}\\ 
  \texttt{\{wwgoing, houhanxu\}@163.com} \quad \texttt{chanceycn@gmail.com}\\ \texttt{daizhongxiang@cuhk.edu.cn} \quad 
  \texttt{\{siyangao,shuanqiu,linqi.song\}@cityu.edu.hk} \\
}

\begin{document}
\maketitle
\begin{abstract}

Although most of the automated theorem-proving approaches depend on formal proof systems, informal theorem proving can align better with large language models' (LLMs) strength in natural language processing. In this work, we identify a primary bottleneck in informal theorem proving as a lack of insight, namely the difficulty of recognizing the core techniques required to solve complex problems.
To address this, we propose \texttt{DeepInsight}, a unified training framework designed to cultivate this essential reasoning skill and enable LLMs to perform insightful reasoning. Our framework consists of three components: (1) \texttt{DeepInsightTheorem}, a hierarchical dataset that structures informal proofs by explicitly extracting core techniques and proof sketches alongside the final proof; (2) a Progressive Multi-Stage SFT strategy that mimics the human learning process, teaching the model proof writing, planning, and insight identification; and (3) \texttt{InsightPO}, a policy optimization method that assigns structured rewards over this insight hierarchy. Our experiments on challenging mathematical benchmarks demonstrate that this insight-aware generation strategy significantly outperforms baselines. These results demonstrate that teaching models to identify and apply core techniques can substantially improve their mathematical reasoning.

\end{abstract}

% Comment them for submission
\renewcommand{\thefootnote}{$*$}
\footnotetext{Equal contribution.}
\renewcommand{\thefootnote}{$\dag$}
\footnotetext{Corresponding author.}

\renewcommand{\thefootnote}{\arabic{footnote}}

\section{Introduction}

Automated theorem proving (ATP) has long been a central goal in the field of artificial intelligence, serving as a key benchmark for evaluating machine reasoning. Recent progress in Large Language Models (LLMs) has greatly changed the field of ATP.  Most previous research has attempted to solve this problem by combining LLMs with formal proof engines like Lean, Coq, and Isabelle \cite{zheng2022minif2f,fimo,putnam} or by using specialized languages \citep{napro}. In contrast, informal theorem proving aims to generate proofs using natural language and standard mathematical notation, often formatted in LaTeX. This setting aligns well with the strengths of modern LLMs. 

% Automated theorem proving (ATP) is a long-standing benchmark for machine reasoning. While most recent LLM-based ATP systems rely on formal proof engines such as Lean, Coq, and Isabelle \cite{zheng2022minif2f,fimo,putnam} or specialized languages \citep{napro}, informal theorem proving asks models to produce human-readable mathematical proofs in natural language and LaTeX. This setting is closer to how mathematicians communicate proofs and is well aligned with LLMs' strengths in language understanding, long-context generation, and mathematical exposition.

% However, only a few works on the informal theorem proving, such as \citet{napro} and \citet{zhang2025deeptheorem}, the area stays highly underexplored. Most of the works have focused on the framework construction and never attend to the proof generation mechanism and the essential bottleneck for LLM informal theorem proving. Additional related work will be discussed in Appendix A.

% However, informal theorem proving remains underexplored \citep{napro,zhang2025deeptheorem}. Existing work mainly builds evaluation or data frameworks, leaving the proof-generation mechanism and its key bottlenecks less studied. In particular, simply fine-tuning on question-proof pairs often encourages models to imitate surface proof patterns, while difficult proof problems require discovering the hidden idea that determines the solution path. Additional related work is discussed in Appendix \ref{sec:related_work}.

However, only a limited number of studies have investigated informal theorem proving, including \citet{napro} and \citet{zhang2025deeptheorem}, and the area remains highly underexplored. Most existing research has focused on framework construction, while paying little attention to the proof generation mechanism and the key bottlenecks of LLM-based informal theorem proving. Additional related work will be discussed in Appendix \ref{sec:related_work}.

Inspired by how human experts prove theorems, we argue that informal theorem proving would require first forming a big-picture view of the proof before eventually completing the full proof. We refer to this early-stage, high-level cognitive act of identifying a set of pivotal ideas or essential technical tools as \textit{insight}. Such pivotal ideas or technical tools identified in this process are termed \textit{core techniques}, typically involving the substantive theoretical machinery, e.g., a specific lemma, theorem, or corollary, rather than merely the elementary routine logical steps. In this paper, we emphasize the importance of insight and core techniques in informal mathematical proofs. Specifically, we identify that the primary bottleneck in proof generation resides in the recognition of these core techniques. We further demonstrate that to improve a model's mathematical reasoning ability, it is feasible and effective to adopt a principled two-phase procedure: \textbf{(1)} extract core techniques by reviewing the proof from a training corpus (the acquisition phase); \textbf{(2)} develop insightful reasoning for a specific problem by identifying the underlying core techniques and then constructing proofs based on them during inference (the application phase).

Empirically, to achieve this, we propose \texttt{DeepInsight}, a unified training framework that integrates data construction, progressive SFT, and reinforcement learning. Our work first introduces \texttt{DeepInsightTheorem}, a newly constructed hierarchical dataset that substantially extends the prior DeepTheorem framework \citep{zhang2025deeptheorem}, with the extraction of core techniques. We then propose a novel progressive multi-stage training strategy, which is carefully designed to emulate key aspects of the human learning process, thereby training LLM for insightful reasoning. Finally, we introduce \texttt{InsightPO}, a policy optimization algorithm that uses this hierarchical structure to provide rewards for core technique identification, proof sketching, and proof completion.

Through extensive experiments, we demonstrate that this insight-aware method significantly outperforms standard baselines. These results confirm that explicitly training the model to acquire proof insight by extracting core techniques enables it to move beyond mere text imitation toward more faithful, insight-driven mathematical reasoning. The primary contributions of our work are summarized as follows:

% Experiments on FIMO, PutnamBench, and HMMT show consistent gains over standard SFT, structured-prompting baselines, and strong open-source theorem-proving models. Because both reinforcement learning and evaluation use LLM-based verification, we also introduce a stratified human-audit protocol to calibrate verifier scores and inspect high-disagreement cases. Our contributions are summarized as follows:

% The pursuit of automated mathematical reasoning stands as a seminal challenge in artificial intelligence (Jiang  Zhou, 2023). Research in this domain has broadly bifurcated into two streams. The first, and perhaps more predominant, focuses on formal theorem proving, where models interact with proof assistants like Lean or Isabelle to generate machine-checkable derivations (Chen et al., 2024). While this approach ensures logical rigor, it often abstracts away the rich, explanatory reasoning expressed in natural language. The second stream targets informal proof generation, aiming to produce human-readable, step-by-step natural language explanations—a task that aligns with the communicative and pedagogical essence of mathematics and fully leverages the core competencies of large language models (LLMs) (Huang et al., 2023). Our work situates itself within this latter category, seeking to advance how LLMs reason and communicate mathematically.
\begin{figure*}
    \centering
    \includegraphics[width=0.8\linewidth]{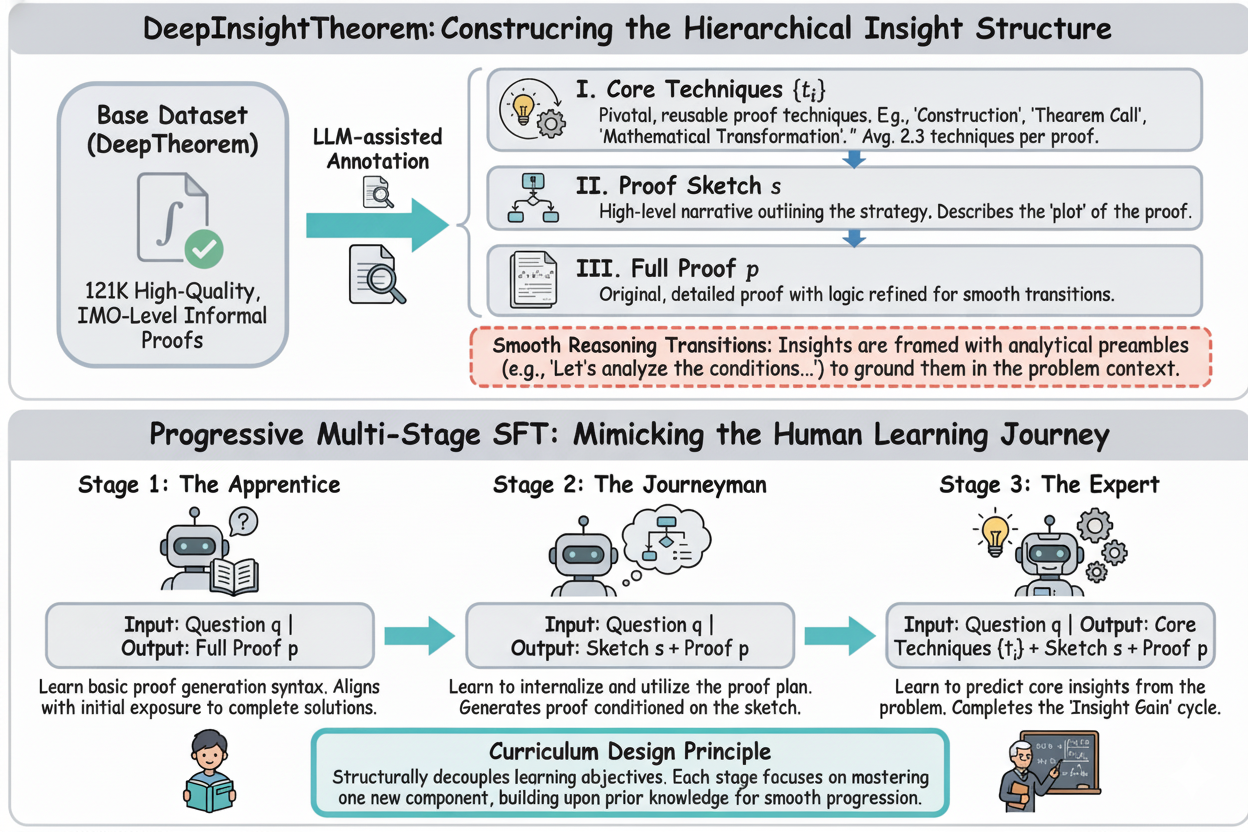}
    \caption{The top figure illustrates  the construction of \texttt{DeepInsightTheorem}, and the bottom figure depicts our progressive multi-stage SFT procedure.}
\label{fig1}
\vspace{-0.5cm}
\end{figure*}

\begin{itemize}[itemsep=1pt, parsep=2pt, leftmargin=10pt, topsep=2pt]

\item We identify core technique recognition as a key bottleneck in informal theorem proving and formalize it through the notion of mathematical insight.
\item We construct \texttt{DeepInsightTheorem}, a hierarchical dataset with core-technique, proof-sketch, and the full proof to facilitate model post-training for informal theorem proving.
\item We propose a novel training pipeline,  \texttt{DeepInsight}, tailored to the hierarchical structure of \texttt{DeepInsightTheorem}, consisting of a progressive multi-stage SFT and \texttt{InsightPO}, a new policy optimization method with structured verifier rewards capturing insight correctness.
\item We evaluate our framework on challenging benchmarks and against strong baselines for mathematical reasoning, and observe clear and consistent improvements over prior methods.

% \item We demonstrate the importance of core techniques in informal theorem proving, showing that models' high perplexity in informal theorem proving largely arises from difficulties in recognizing and applying the core techniques.
    
% \item We introduce the notion of mathematical insight and propose an insight-guided reasoning paradigm, which can improve mathematical reasoning when paired with a well-designed method.

% \item We construct a novel hierarchical dataset, named \texttt{DeepInsightTheorem}, by explicitly extracting core techniques from base datasets, which facilitates both technique review and insight-driven reasoning.

% \item We design a progressive, multi-stage training scheme that effectively leverages the hierarchical structure of \texttt{DeepInsightTheorem} to enhance informal theorem proving.

% \item We propose \texttt{InsightPO}, a GRPO-style reinforcement learning algorithm, forming the complete \texttt{DeepInsight} framework that achieves state-of-the-art informal theorem-proving performance across all tested model scales.

% \item We evaluate our framework across challenging benchmarks, structured prompting baselines, and a human-audited evaluation protocol.
\end{itemize}

% 4. {\bf An Insight Dataset of Mathematical Proof}: We release a new mathematical proof dataset based on standard datasets to assist with deeper research into mathematical proofs and reasoning.

\section{Preliminary}

% \subsection{Foundations}
We consider the context of informal mathematical proof generation. 
    Let $\mathcal{V}$ be the vocabulary, which is a finite discrete set. The text sequence space $\mathcal{S}$ is defined as the union of the product spaces of any finite power 
     $\mathcal{S} \triangleq \bigcup_{l=0}^{\infty} \mathcal{V}^l.$

We can then denote the space of theorems as a subspace $\mathcal{Q}$ of $\mathcal{S}$ which contains all well-defined math problems for theorem-proving. For any $q\in \mathcal{Q}$, we denote the space of proofs with respect to $q$ as $\mathcal{P}(q) \subset \mathcal{S} $  that contains all valid proofs of $q$. 

The LLM $M_{\theta}$ is an auto-regressive network which will predict the next token's distribution over the vocabulary when it is working.
To train a generation policy $\pi_{\theta}$ to be a theorem-prover, the training datasets are collected from many resources of math corpus and commonly formed as theorem-proof pairs $(q, p)$.  For Supervised Fine-Tuning (SFT) on such a dataset, the objective is to solve $\max_\theta ~ \mathbb{E}_{q \sim \mathcal{D}, p\sim\mathcal{D}_{ P(q)  } } \pi_\theta(p | q)$,
where $\mathcal{D}$ is the distribution of the math problems in the training set and $\mathcal{D}_{ P(q)  }$ is the distribution over the valid proofs of each question in the training set.

\subsection{Notion of Insight} \label{2.2}

For a mathematical proof question, \textit{insight} is a cognitive act of identifying a set of pivotal ideas (e.g., ``apply the Pigeonhole Principle'', ``utilize a specific invariant xxx'' ) through the given conditions to capture the essence of the solution. We refer to such pivotal ideas as \textit{core techniques} in this paper. Insight is not the proof itself, but instead refers to a preliminary, high-level perception and foresight of the core techniques required to construct the proof.
% Insight is not the proof itself, but instead refers to a preliminary, high-level perception and foresight of the core techniques required to construct the proof. In this sense, insight is a cognitive act of identifying a set of pivotal ideas (e.g., ``apply the Pigeonhole Principle'', ``utilize a specific invariant xxx'' ) through the given conditions to capture the essence of the solution. We refer to such ideas as \textbf{core techniques} in this paper.
% {\color{red}(We refer to such ideas (insight) as core techniques? what is the difference between insight and core techniques?)} 
Generally, we group these techniques into three main categories that cover the most common and essential technical patterns in math proofs:  
\begin{itemize}[itemsep=1pt, parsep=2pt, leftmargin=10pt, topsep=2pt]
    \item {\bf Construction}: Introducing auxiliary objects, e.g., ``Define the sequence \(x_{n+1} = U x_n\) for \(n \geq 0\). This Picard iteration constructs the sequence \(x_0, x_1 = U x_0, x_2 = U^2 x_0, \dots\)''.
    \item {\bf Theorem Call}: Invoking a known lemma, theorem, or any existing result, e.g. ``By using Cantor's Theorem.''
    \item {\bf Mathematical Transformation}: Performing a reformulation that recasts the problem in a new framework, e.g., ``Define a topology $\tau$ on the set of integers $\mathbb{Z}$..., shifting the language and tools entirely from number theory to topology.'' 
\end{itemize}
Formally, we denote the space of techniques as $\mathcal{T} \subset \mathcal{S}$  that contains all possible techniques in those three classes. Generally, we can then write each proof $p$ of $q$ as an ordered sequence
\begin{equation}\label{p}
    p = (r_1, t_1, r_2, t_2, \cdots, r_k), 
\end{equation}
where $t_i \in \mathcal{T}$, $r_i$ denote an elementary reasoning statement between successive techniques, and $k$ is the total number of techniques in the proof.

\section{Motivating Example: Insightfulness Evaluation of Off-the-Shelf LLMs} \label{2.3}

Expert mathematicians can often quickly form a big-picture view of a proof when facing a novel problem, drawing on insights cultivated through experience.
% In contrast, general-purpose LLMs often fail to form such high-level ideas at the outset of proof generation.
In contrast, general-purpose LLMs often fail to reliably form such high-level ideas at the outset of proof generation, and are not as adept as experts at identifying the pivotal technique early.

We evaluated several powerful commercial LLMs with a non-thinking mode\footnote{Long chain-of-thought reasoning often make the model first go over the whole proof and then summarize the core techniques, which is against the thinking pattern of insight. Therefore, it is more appropriate to disable the thinking mode.}, including Gemini 2.5 Flash \citep{gemini} and DeepSeek-R1 \citep{r1}, by prompting them to provide insights for math competition problems randomly sampled from Putnam and FIMO datasets (See Appendix \ref{prompts} for detailed prompts). We then use o3-mini to review and evaluate the insights generated by these models from the dimensions of depth and completeness (Appendix \ref{exp2.3}). Here, depth indicates the insight is specific and technically accurate, rather than a generic, high-level statement or a vague direction, while completeness measures whether all core techniques required for the solution are identified. Our results show that, in terms of depth, both models largely failed to provide specific and accurate insights, instead producing shallow guesses or vague hints at some directions. In terms of completeness, they always see just one or two low-hanging ideas while missing more central techniques.

\begin{figure*}[t]
\centering
  \includegraphics[width=0.75\linewidth, height=6cm]{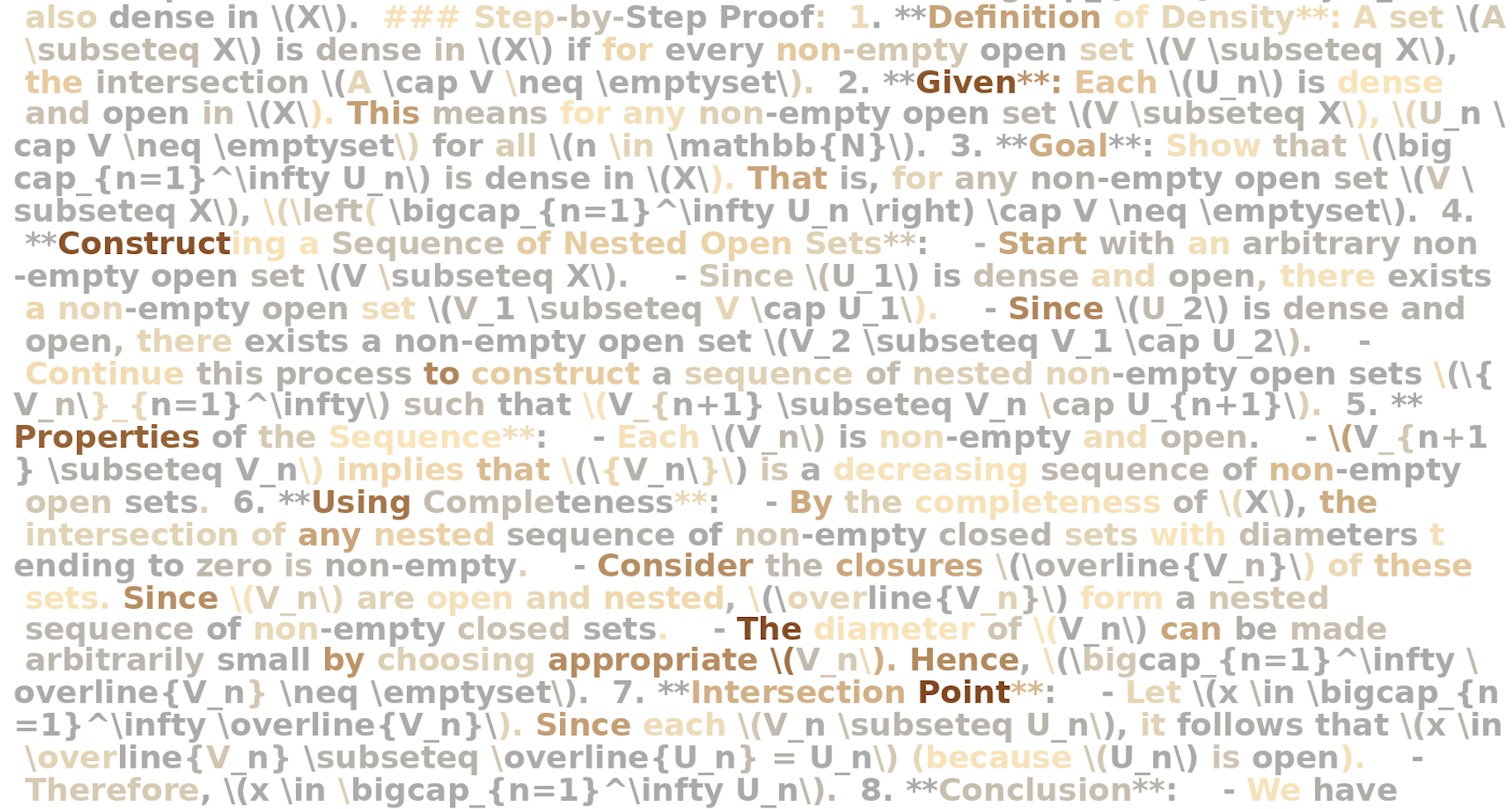}
  \caption{The generated proof of the Baire Category Theorem generated by Qwen2.5-14B-Instruct. From light grey to deep warm color, the token entropy increases. Most tokens are of low entropy, composing basic mathematical reasoning. Tokens with a spiking entropy indicate core techniques.}
  % \label{fig:experiments}
  \label{fig2}
\end{figure*}

\vspace{-0.15cm}

\section{Methodology} \label{3}

\vspace{-0.15cm}

% \vspace{-0.5cm}

\subsection{Challenge of Identifying Core Techniques} \label{3.1}

\vspace{-0.1cm}

The main body of a proof is largely composed of some basic, easy-to-learn mathematical reasoning statements. For a theorem-proving LLM denoted by $\pi_{\text{TP}}$ generating the proof for a problem, the uncertainty remains low until a core technique needs to be recognized. This likely arises for two main reasons. First, core techniques are intrinsically harder to identify. Second, since basic reasoning steps constitute a large fraction of the proof, LLM training may overly emphasize these basic reasoning details rather than core technique recognition.

% This fact is illustrated in Figure \ref{fig2}.

% When we train LLM on the plain theorem-proof pairs, where we denote the trained LLM as $\pi_{\text{TP}}$, it  will hence mainly focus on those basic reasoning and will be hard to recognize the core techniques. 

 % As we can see the generated proof of the well-known Baire Category Theorem in General Topology by Qwen2.5-14B-Instruct \citep{qwen} from Figure \ref{fig2}, tokens with a soaring entropy are always indicating core techniques:

As illustrated in Figure~\ref{fig2}, in the proof generated by Qwen2.5-14B-Instruct~\citep{qwen} for the well-known Baire Category Theorem in General Topology, tokens with spiking entropy consistently signal the presence of core techniques:

\vspace{3pt}
\noindent     \underline{\emph{Construction}}: ``\texttt{Constructing a Sequence of ...}'' to construct a sequence of nested open sets.

\vspace{3pt}
\noindent     \underline{\emph{Theorem call}}: ``\texttt{Using completeness...}'' to apply the completeness of the ground space.

\vspace{3pt}
\noindent     \underline{\emph{Mathematical transformation}}: ``\texttt{Consider the closures...}'' to transfer the issue of open sets to their closures. 

Note that the above three techniques are defined in Section \ref{2.2}. The rest of the proof is based on basic elementary reasoning statements that connect these core techniques, making the overall argument complete and smooth.

As discussed above, the model $\pi_{\text{TP}}$ struggles to decide which core technique to apply, as evidenced by the token-level entropy spikes at positions where a core technique is introduced. To further formalize this, we assume 
\begin{equation} \label{etp}
    \pi_{\text{TP}}(t_i |u_i,  q) < \delta
\end{equation}
% {\color{red} $\pi_{\text{TP}}$ is not defined. Should it be $\pi_{\theta}$ or $\pi^*$?}
for some small constant $\delta >0$ and any $t_i \in \mathcal{T}$, where $u_i := (r_1,t_1,\cdots,r_i)$ denotes the prefix before $t_i$ and $i$ indexes the $i$-th technique in a proof of form \eqref{p}. This reflects the difficulty or the high uncertainty of identifying correct techniques. Consequently, we have the probability of directly predicting a core technique $t_i$ from the question $q$ bounded by $\delta$: $    \pi_{\text{TP}}(t_i | q)  = \sum_{u_i} \pi_{\text{TP}}(t_i |u_i,  q) \pi_{\text{TP}}(u_i |  q) < \delta.$
% \begin{equation*} 
%     \pi_{\text{TP}}(t_i | q)  = \sum_{u_i} \pi_{\text{TP}}(t_i |u_i,  q) \pi_{\text{TP}}(u_i |  q) < \delta.
% \end{equation*}
% where $s$ is over the possible prefix. 
Similarly, by recursively applying \eqref{etp}, for the probability  
% {\color{red} what is  $\pi_{\text{TP}}(p|q)$?}
of generating a valid full proof $p$ for $q$ via $\pi_{\text{TP}}$, we have 
$\pi_{\text{TP}}(p|q) \leq \delta^k,$ where $k$ is the total number of techniques appearing in the proof. This indicates that the ability to complete a proof is restricted by the uncertainty. A detailed derivation is in Appendix \ref{a1}.
We can interpret the value $\pi_{\text{TP}}(t_i | q)$ as the \emph{insightfulness} of $\pi_{\text{TP}}$ in identifying technique $t_i\in \mathcal{T}$. A higher value indicates stronger insight in recognizing $t_i$ as the appropriate technique to apply.

% {\color{red}(call a policy as insightfulness?)}

Therefore, the successful realization of core techniques is crucial for generating valid proofs in theorem proving. Core techniques also serve as more appropriate memory anchors for key pattern recognition in complex mathematical reasoning. This prompts a necessity of enhanced learning to bridge the gap between the question and the core techniques required to solve it.

\subsection{Insight Acquisition and Application} \label{3.2}
It is critical to figure out the core techniques for the proof of a specific question if we want an LLM to grasp the essence of such a proof. After identifying the relevant techniques, the remaining challenge is to integrate these core techniques into a coherent solution, which relies primarily on the fundamental mathematical reasoning capability. 

This reflects a shift in the underlying reasoning pattern for an LLM. In human learning, when presented with a problem and its corresponding proof, beginners typically need to carefully study and review the entire solution to gradually identify and understand the core techniques involved. Subsequently, when faced with similar problems later, they will first form an initial idea about the core techniques and then use the intuition to construct the proof.
% For an LLM $\pi_{\theta}$ (parameterized by $\theta$) to be trained, by the total probability decomposition, we have
% \begin{equation}\label{get}
%     % \pi_{\theta}(t_i | q) = \sum_{p} \pi_{\theta}(t_i |p, q)\pi_{\theta}(p | q).
% \end{equation}
% % Unlike \eqref{tech}, here we consider generating an entire proof $p$.
% Here $\pi_{\theta}(p | q)$ is the standard theorem-to-proof conditional distribution that the model is expected to learn by training on plain theorem-proof pairs, while $\pi_{\theta}(t_i |p, q)$ captures how well $\pi_{\theta}$ can distill a core technique $t_i$ from the proof $p$ of $q$. We interpret the value $\pi_{\theta}(t_i | p, q)$ as $\pi_{\theta}$'s \emph{acquisition strength} in extracting technique $t_i \in \mathcal{T}$ from $p$.

% {\color{red}(call a policy as acquisition?)}

% From \eqref{tech} and \eqref{get},  we know that what we still need for insightfulness is 
% the acquisition. 

Our framework comprises the following two  complementary cognitive processes, mirroring how humans consolidate and review learning materials: 
\begin{itemize}[itemsep=1pt, parsep=2pt, leftmargin=10pt, topsep=2pt]
    \item \textbf{Acquisition phase}: extract core techniques by reviewing proofs in a training corpus;
    \item \textbf{Application phase}: develop insightful reasoning by first predicting the core techniques and then generating proofs by applying these techniques.
\end{itemize}

We consider the following generation process for a proof $p$ via a policy $\pi_{\theta}$ parameterized by $\theta$:
\begin{equation}\label{insight}
    \pi_{\theta}( t, p | q) = \pi_{\theta}(t | q)\pi_{\theta}(p | t, q), 
\end{equation}
where $t = (t_1 ,\cdots, t_k )$. In \eqref{insight}, we first identify the required core techniques \(t\) via $\pi_{\theta}(t | q)$ before generating the proof \(p\). The proof \(p\) is then generated conditioned on the identified techniques \(t\) via $\pi_{\theta}(p | t, q)$. Consequently, the performance of \(\pi_{\theta}\) depends critically on how well the first factor \(\pi_{\theta}(t \mid q)\) is learned. In the following sections, we construct a well-curated dataset and propose a novel multi-stage method to optimize \(\theta\) such that both \(\pi_{\theta}(t \mid q)\) and \(\pi_{\theta}(p \mid t, q)\) are trained effectively, which corresponds to the acquisition phase. Subsequently, during the application phase, we apply \eqref{insight} for proof generation.

Furthermore, we can consider a more refined generation process:
\begin{equation}\label{insight-sketch}
\pi_{\theta}(t, s, p | q)
=
\pi_{\theta}(t | q)
\pi_{\theta}(s | t, q)
\pi_{\theta}(p | s, t, q),
\end{equation}
where \(s\) denotes a proof sketch generated conditioned on the identified core techniques $t$. In this scenario, the proof \(p\) is generated conditioned on both the core techniques \(t\) and the sketch \(s\). This hierarchical decomposition will make the model first commit to a high-level thinking structure before generating a full proof, reflecting a natural human problem-solving process. Furthermore, we follow the dual phases of insight acquisition and application for effectively generating mathematical proofs. Next, we show how to instantiate these processes in practice via more concrete procedures.

\section{Training an Insightful LLM}

\subsection{Dataset Construction} \label{4.1}

% Our hierarchical dataset is engineered to explicitly instantiate the review process, making the core techniques explicitly learnable. To ensure high quality and sufficient challenge, we construct our data based on the DeepTheorem \citep{zhang2025deeptheorem} dataset, a recently introduced, large-scale resource for informal mathematical theorem proving. 

We construct a hierarchical dataset to train an insightful LLM that can explicitly identify and subsequently generate the core techniques to guide the proof of a given problem. 
We design the dataset to provide a supervision signal that mirrors human review and consolidation. To ensure high quality and sufficient difficulty, we build our data based on the DeepTheorem \citep{zhang2025deeptheorem} dataset, a recently introduced, large-scale resource for informal mathematical theorem proving.

% To ensure both high quality and sufficient difficulty, we build our dataset on DeepTheorem \citep{zhang2025deeptheorem}, a recently introduced large-scale resource for informal mathematical theorem proving.

% \vspace{5pt}
% \noindent{\bf Base Dataset.} The DeepTheorem dataset provides a robust foundation consisting of 121K high-quality, informal theorem-proof pairs at IMO-level difficulty. Its construction involves a rigorous pipeline including sourcing from diverse corpora and implementing strict decontamination against major benchmarks (e.g., MATH \citep{math}, AIME \citep{aime}, miniF2F \citep{zheng2022minif2f}). Since DeepTheorem already contains many common datasets used for theorem-proving training, we believe the experiments based on it can efficiently and effectively establish a generalized average performance result for our methods.   One can refer to \citet{zhang2025deeptheorem} for more details about the dataset.

\vspace{5pt}
\noindent{\bf Base Dataset.} DeepTheorem \citep{zhang2025deeptheorem} provides a robust foundation of 121K high-quality informal theorem-proof pairs at roughly IMO-level difficulty. It is constructed through a rigorous pipeline, including collection from diverse corpora and strict decontamination against major benchmarks (e.g., MATH \citep{math}, AIME \citep{aime}, miniF2F \citep{zheng2022minif2f}). Since DeepTheorem already covers many commonly used datasets for theorem-proving training and is built through a rigorous, high-quality construction pipeline, using it as the base dataset allows our subsequent hierarchical annotation to inherit these desirable properties, resulting in a curated dataset with strong overall quality. We refer readers to \citet{zhang2025deeptheorem} for further details.

\vspace{5pt}
\noindent{\bf DeepInsightTheorem.} Since LLMs struggle to extract core techniques on their own (Section \ref{3.1}), we augment model training with a richer supervision signal by explicitly providing core techniques. Building upon DeepTheorem's theorem-proof data pairs $(q,p)$, we perform additional annotations to transform each proof into a hierarchical representation as follows:
% $$ (            q, \quad \underbrace{t_1, \dots,t_m}_{\text{core techs}} \quad, \underbrace{\quad s \quad}_{\text{sketch}}, \underbrace{\quad p \quad}_{\text{proof}}      )     $$
% \[
% \Bigl(
%   q,\;
%   \underbrace{t_1,\dots,t_m}_{\text{core techs}},\;
%   \underbrace{\;\;s\;\;}_{\text{sketch}},\;
%   \underbrace{\;\;p\;\;}_{\text{proof}}
% \Bigr)
% \]
\[
\bigl(
  q,\;
  \underbrace{(t_1,\ldots,t_m)}_{\text{core techs}},\;
  \underbrace{\makebox[2.9em][c]{$s$}}_{\text{sketch}},\;
  \underbrace{\makebox[2.2em][c]{$p$}}_{\text{proof}}
\bigr).
\]
This process involves a meticulous, LLM-assisted analysis to extract deep information from proof $p$, following a dedicated prompt to design each component as follows:   
\begin{itemize}[itemsep=1pt, parsep=2pt, leftmargin=10pt]
\item {\bf Core Techniques $t_i$.} Instead of only listing the core techniques, we first include a guiding statement as “\texttt{Let's analyze the conditions...}”, then those core techniques are introduced with heuristic language, for example: “\texttt{The condition ... tells us to construct an auxiliary function...} ” or “\texttt{... suggests using xxx Theorem.}”. The core techniques are themselves finally summarized at the end of this component according to those three main classes. 
\item {\bf Proof Sketch $s$.} We insert an intermediate component between the core techniques and the full proof. This sketch outlines a proof strategy based on the identified core techniques, bridging high-level insights and low-level derivations to encourage a smooth and coherent reasoning flow. 
% {\color{red} What is the difference between $s$ here and $s_i$ in (2)? Should we use different notations?}

\item 
{\bf  Proof $p$.} The original, detailed proof from the base dataset serves as the ground-truth instantiation. 

\end{itemize}
We refer to the resulting dataset as \textbf{\texttt{DeepInsight-\allowbreak Theorem}}, highlighting the insight-driven hierarchical structure derived from DeepTheorem. See the top of Figure \ref{fig1} for an overview, and Figure \ref{figdata} for a concrete example.

\vspace{5pt}
\noindent{\bf Chain-of-Thought in Insight}. A key design consideration in the construction of \texttt{DeepInsightTheorem} is to avoid presenting the core techniques as an isolated list. Instead, we frame insight generation with a short, analytical preamble that serves as a localized micro chain-of-thought for analyzing the question.
% {\color{red} (condition analysis from the question? not sure what it means)}
% Throughout reasoning, the LLM may hence leads a more accurate insight prediction compared to directly outputting a  list of techniques.
This encourages a more accurate insight prediction than directly outputting a list of techniques.

\vspace{5pt}
% \noindent{\bf  Review-Application Logical Transformation.} There exists a subtle but crucial logical shift between the review and application dual processes, even though the main contents throughout these two processes are basically the same. The former is backward-looking: “ \texttt{Analyzing this proof, we see that xxx is the core tech.}”, while the latter is forward-looking: “ \texttt{Given these problem conditions, xxx might be a core tech.}” These two processes by our design are integrated into the LLM-assisted data annotation. Therefore we need to carefully tune the prompt for the assistant LLM to ensure the smoothness of the transformation. The specific prompts used are detailed in Appendix \ref{a2}.

\noindent{\bf  Acquisition-Application Transition.} There exists a subtle but crucial logical shift between acquisition and application phases, even though the two processes are basically similar. The former is backward-looking: ``\texttt{Analyzing this proof, we see that xxx is the core tech.}'', while the latter is forward-looking: ``\texttt{Given these problem conditions, xxx might be a core tech.}'' These two processes by our design are integrated into our LLM-assisted data annotations.
% {\color{red}(The reviewers would rather see how they are integrated.)} 
Therefore, we need to carefully tune the prompt for the  LLM  to ensure the smoothness of the transformation. The specific prompt shows how the two processes are integrated, as for each problem, this prompt on one hand encourages the LLM to analyze the corresponding proof through summarization and extract the core techniques as required, and on the other hand shifts perspective to play the role of an expert encountering the problem for the first time, integrating these core techniques into the thought process of analyzing the problem. The prompts used are detailed in Appendix \ref{prompts}. 

% {\color{red} (Appendix C is a proof, not prompt. This part is not clear enough)}
% (for reference...) Both perspectives are incorporated into our LLM-assisted annotations, which requires careful prompt design to ensure a smooth transition between the two. The specific prompts used are detailed in Appendix C.

% {\bf Maximizing Data Utility}. A fundamental principle of DeepInsightTheorem is its self-contained nature. The creation of the hierarchical structure $(q, \{t_i\}, s,  p)$ is achieved solely through the analysis of the information already embedded within the original proofs.  By doing so, we increase the informational density  of each training example, offering a powerful pathway to improve data efficiency when scaling high-quality mathematical proof data.

\begin{figure}
    \centering
    \includegraphics[width=0.6\linewidth]{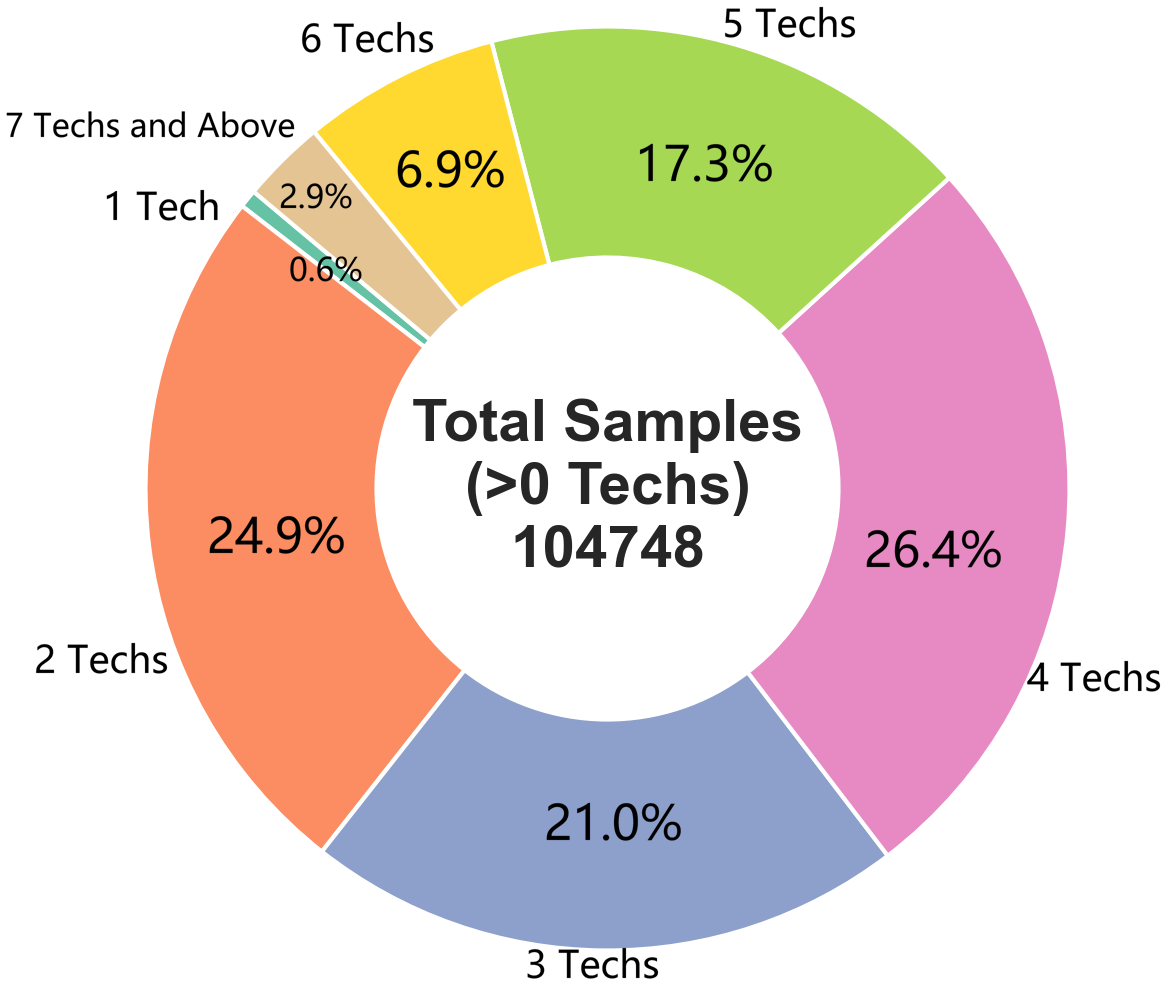}
    \caption{Distribution of core technique counts}
    \label{techno}

    \vspace{-0.2cm}
    
\end{figure}

\vspace{5pt}
\noindent{\bf Key Statistics}. Our curated dataset contains approximately 100K problems. On average, each proof involves 3.6 core techniques, with more complex problems frequently combining four or more, as illustrated in Figure \ref{techno}. A distribution of core technique counts is shown in Figure \ref{bar2}.

\subsection{Progressive Multi-Stage SFT}\label{sft}

% The other process consists in the application of the core techniques summarized above (Section \ref{3.2}). The LLM can already have a hierarchical output due to the corresponding structure of DeepInsightTheorem and generate the core technique analysis first (Section \ref{4.1}). It remains to train LLM appropriately such that the two processes {\color{red}(What two processes? Do not always use `process' to refer something important, which is quite unclear to readers)} in \eqref{insight} both work well. In principle, the decomposition \eqref{get} and \eqref{insight} have already implied a two-stage training paradigm, i.e., the acquisition requires LLM to train on the original theorem-proof pairs $(q, p)$ and the application to continue training on insight induced data.

The next step concerns applying the core techniques summarized above. Thanks to the hierarchical structure of \texttt{DeepInsightTheorem}, the model can already be trained to produce hierarchical outputs, generating a core technique analysis first (Section \ref{4.1}).
% What remains is to train the LLM appropriately so that both processes 
% {\color{red}(What two processes? Do not always use `process' to refer something important, which is quite unclear to readers)} 
% in \eqref{insight} are reliably learned.
Then the decompositions in  \eqref{insight} naturally suggest a two-phase training paradigm: the acquisition phase trains the LLM on the original theorem-proof pairs $(q,p)$, while the application phase continues training on the insight-induced data.

We empirically demonstrate the necessity of this two-phase design by comparing training solely on \texttt{DeepInsightTheorem} against the baseline. Results for Qwen2.5-7B and Llama3-8B are reported in Table \ref{2stage}.   
The comparison indicates that directly training on \texttt{DeepInsightTheorem} does not necessarily allow the LLM to benefit from the insight structure. 
% One plausible reason is that both \eqref{get} and \eqref{insight} implicitly assume a strong foundation learned from theorem--proof data, i.e., sufficiently large $\pi_{\theta}(p \mid q)$ and $\pi_{\theta}(r_i \mid r^{\le i-1}, t, q)$, respectively. 
One of the reasons is that a higher value of  \eqref{insight}   assumes a training on theorem-proof data to enlarge $\pi_{\theta}(p | q)$ and $\pi_{\theta}(r_i | r^{\le i-1}, t, q)$ respectively. 
% {\color{red}(Not clear. Not sure how you get this claim from \eqref{get} and \eqref{insight}.)}
Without first training on theorem-proof pairs, the model may lack the basic mathematical reasoning and proof-writing competence required to exploit technique-level guidance. From a learning perspective, this can also be viewed as a mismatch with the natural progression of human learning: teaching a novice the expert's high-level reasoning pattern before solidifying foundational proof-writing skills can create a substantial comprehension gap.

% It shows that LLM being directly trained on DeepInsightTheorem may not be able to gain benefits from the insight structure. 
% One of the reasons is that  higher value of  \eqref{get}
% and \eqref{insight} both  assume a training on theorem-proof data to enlarge $\pi_{\theta}(p | q)$ and $\pi_{\theta}(r_i | r^{\le i-1}, t, q)$ respectively. Missing the theorem-proof pair training it may not be able to have a good basic mathematical reasoning capability. On the other hand,  this can also be viewed as a misalignment with the natural progression of learning: It may create a comprehension gap to directly teach a novice learner   the expert's complex thinking pattern without first solidifying foundational proof-writing skills.

 % While our hierarchical dataset explicitly structures the expert reasoning chain, we empirically observed that performing standard Supervised Fine-Tuning (SFT) directly on this complete structure often yields performance similar to, or sometimes worse than, training on the base $(q,p)$ pairs. We hypothesize that this stems from a \emph{ joint learning dilemma}. When trained on the full hierarchy, the model's parameters struggle to concurrently develop distinct competencies for abstraction, planning, and execution over limited training epochs, leading to suboptimal convergence where none of the components is mastered. 
 
Inspired this analysis, we adopt a multi-stage training strategy for fine-tuning LLMs. Specifically, we propose a three-stage curriculum that progresses from direct proof generation to sketch-conditioned generation and finally technique-guided reasoning, which we refer to as \emph{Apprentice}, \emph{Journeyman}, and \emph{Expert}. Below we outline the main stages of our progressive multi-stage SFT, each conducted on the same dataset but using different components of \texttt{DeepInsightTheorem}:

% \begin{table}
%   \centering
%   \renewcommand{\arraystretch}{1.1}
%   \begin{tabular}{lccc}
%     \toprule
%     \textbf{Methods} & \textbf{FIMO} & \textbf{Putnam} & \textbf{HMMT}  \\
%     \addlinespace[2pt]
%     \hline
%     Base     &     & &       \\
%     Our     &      & &     \\
%     \bottomrule
%   \end{tabular}
%     \begin{tabular}{lccc}
%     \toprule
%     \textbf{Methods} & \textbf{FIMO} & \textbf{Putnam} & \textbf{HMMT}  \\
%     \addlinespace[2pt]
%     \hline
%     Base     &     & &       \\
%     Our     &      & &     \\
%     \bottomrule
%   \end{tabular}
%     \caption{Caption}
%     \label{tab:placeholder}
% \end{table}

\begin{table}
  \centering
  \renewcommand{\arraystretch}{1.1}

    \begin{tabular}{lccc}
      \toprule
      \textbf{Methods} & \textbf{FIMO} & \textbf{Putnam} & \textbf{HMMT}  \\
      \addlinespace[2pt]
      \midrule
      Qwen2.5-7B     &   15.73  & 37.01   &  12.59     \\
      Llama3-8B     &  12.50    & 36.69 &  9.98    \\
      \bottomrule
    \end{tabular}
    \caption{Evaluation results for models solely trained on \texttt{DeepInsightTheorem} }
    % \label{tab:first}

  % \caption{Yea}
  \label{2stage}

  \vspace{-0.2cm}
\end{table}

  \begin{itemize}[itemsep=1pt, parsep=2pt, leftmargin=10pt, topsep=2pt]  

    \item {\bf Stage 1 (Apprentice): $(q, p)$.} 
    % Learn the basic task of proof generation  on the plain theorem-proof pairs. This aligns with initial exposure to complete solutions, reflecting a junior learner referring to the answer proof.
    We first train the model on plain theorem-proof pairs to acquire basic proof-generation ability. This stage corresponds to a novice learner studying complete solutions.
    
    \item {\bf Stage 2 (Journeyman): $(q, s, p)$.} 
    % Learn to generate the proof plan on the theorem-sketch-proof triples. The model must now generate the proof conditioned on the high-level sketch, which begins the abstraction process. It reflects a senior learner's behavior where he can not only copy the proof but know the logic structure by steps.
    We then train the model on theorem-sketch-proof triples, encouraging it to condition proof generation on a high-level sketch. This introduces an intermediate, coarse-to-fine abstraction and reflects a more advanced learner who can follow and internalize the step-wise logical structure rather than merely reproducing surface forms.

    \item  {\bf Stage 3 (Expert): $(q, \{t_i\}, s,  p)$.} 
    % Learn to insightfully predict core techniques from the problem and use them to derive the sketch and proof on the whole hierarchical data. This stage completes the dual processes, teaching the model to map the question directly to the core techniques. It reflects hence the expert's thinking paradigm.
    Finally, we train on the full hierarchical structure, so that the model learns to predict core techniques from the problem and leverage them to derive the sketch and the final proof. This stage explicitly learns a direct mapping from the question to its core techniques, resembling an expert's technique-driven reasoning paradigm.
    
    \end{itemize}

    This multi-stage training strategy is illustrated in Figure~\ref{fig1} bottom. Overall, the progressive multi-stage design structurally decouples the learning objectives, yielding a clear and focused target for each training stage. Each stage's training primarily emphasizes the mapping from the problem to the newly introduced component (e.g., $p$, then $s$, then $\{t_i\}$). Since each mapping is not yet adequately learned by the model, continued training does not cause overfitting.

\subsection{Insight Policy Optimization}\label{InsightPO}

After the progressive multi-stage SFT, the model has learned to generate proofs in the hierarchical format of \texttt{DeepInsightTheorem}. We further propose \texttt{InsightPO}, a GRPO-style reinforcement learning algorithm tailored to this framework. The key idea is to align reinforcement learning with our structured data format: instead of assigning only an outcome reward to the entire proof, the verifier evaluates whether the generated insights, proof sketch, and final proof form a coherent hierarchy, providing a hierarchy-aware reward that reflects both the quality of the identified insight and the validity of the resulting proof. This structured reward is then used in group-relative policy optimization, encouraging the model to improve not only final proof writing but also the intermediate insight and planning behavior that supports difficult theorem proving.
    
    % An illustration can be seen on  Figure \ref{fig1} bottom. Correspondingly, the progressive multi-stage design structurally decouples the learning objectives, providing a clear and focused target for each training stage. In each stage, the primary parameter updates are concentrated on learning the mapping between the problem and the newly introduced component. Since this mapping is not yet adequately learned by the model, continued training would not cause overfitting. 

% Simultaneously, the newly introduced components are logically and smoothly connected to those already learned. After learning on, for instance $(q, p)$ pairs, building upon its understanding of the proof $p$, the model can more readily learn to associate the sketch $s$ with $p$ and correctly complete the proof $p$ in the next stage (learning on $(q, s, p)$).

% The sequential progression (Stages 1 → 2 → 3) simulates the gradual Insight Gain of an expert. The final stage's input-output format becomes the inference-time protocol for the trained model. When presented with a new problem $q'$ at test time, the model is prompted to generate in the \texttt{<tech>...</tech><sketch>...</sketch>}
% \texttt{<proof>...</proof>} format, e.g., the insight content is enclosed inside  \texttt{<tech>...</tech>}.

\begin{table*}[t]\small
\centering
  \renewcommand{\arraystretch}{0.4} % 增加行高
  \setlength{\tabcolsep}{12pt} % 增加列间距
\begin{tabular}{l l c c c c}
\toprule
\textbf{Models} & \textbf{Methods} & \textbf{FIMO} & \textbf{Putnam} & \textbf{HMMT} & \textbf{Avg.} \\
\midrule
\multirow{4}{*}{Qwen2.5-1.5B} & Base       & 12.63 & 26.66 & 12.79  & 17.36  \\
\cmidrule(lr){2-2}
                             & Two-stage  & 13.09 & 30.06 & 13.42 & 18.86 \\
                             & Three-stage & 14.35 & 30.21 & 15.30 & 19.95 \\
\cmidrule(lr){2-2}
                             & \texttt{DeepInsight}& {\bf 17.60} & {\bf 35.50} & {\bf 17.40} & {\bf 23.50} \\
\midrule

\multirow{4}{*}{Qwen2.5-3B} & Base       & 12.92 & 32.94 & 17.04 & 24.30 \\
    \cmidrule(lr){2-2}
                             & Two-stage  & 15.05 & 35.00 & 18.88 & 28.31 \\
                             & Three-stage & 14.99 & 35.94 & 20.21 & 28.71 \\
    \cmidrule(lr){2-2}
                             & \texttt{DeepInsight}& {\bf 21.00} & {\bf 47.60} & {\bf 24.55} & {\bf 31.05} \\
\midrule

\multirow{4}{*}{Qwen2.5-7B} & Base       & 16.89 & 37.76 & 15.52 & 23.39 \\
    \cmidrule(lr){2-2}
                             & Two-stage  & 18.39 & 45.20 & 17.76 & 27.12 \\
                             & Three-stage & 19.31 & 43.35 & 18.49 & 27.05 \\
    \cmidrule(lr){2-2}
                             & \texttt{DeepInsight}& {\bf 27.80} & {\bf 55.50} & {\bf 24.60} & {\bf 35.97} \\
\midrule

\multirow{4}{*}{Llama3.2-1B} & Base       & 6.72 & 16.20 & 4.25 & 9.06 \\
    \cmidrule(lr){2-2}
                             & Two-stage  & 6.91 & 20.78 & 6.76 & 11.48 \\
                             & Three-stage & 8.49 & 21.55 & 8.02 & 12.69 \\
    \cmidrule(lr){2-2}
                             & \texttt{DeepInsight}& {\bf 11.20} & {\bf 25.30} & {\bf 11.00} & {\bf 15.83} \\
\midrule

\multirow{4}{*}{Llama3.2-3B} & Base       & 10.46 & 24.43 & 8.84 & 14.58 \\
    \cmidrule(lr){2-2}
                             & Two-stage  & 11.98 & 24.69 & 11.09 & 15.92 \\
                             & Three-stage & 12.54 & 27.51 & 10.77 & 16.94 \\
    \cmidrule(lr){2-2}
                             & \texttt{DeepInsight}& {\bf 17.65} & {\bf 36.50} & {\bf 14.10} & {\bf 22.75} \\
\midrule

\multirow{4}{*}{Llama3-8B}  & Base       & 15.61 & 38.65 & 13.67 & 22.64 \\
    \cmidrule(lr){2-2}
                             & Two-stage  & 16.70 & 42.76 & 16.51 & 25.32 \\
                             & Three-stage & 15.98 & 41.61 & 16.64 & 24.74 \\
    \cmidrule(lr){2-2}
                             & \texttt{DeepInsight}& {\bf 22.10} & {\bf 50.20} & {\bf 20.30} & {\bf 30.87} \\
\bottomrule
\end{tabular}
\caption{Average evaluation scores of models trained with our methods and the baseline by DeepSeek-R1 and o3-mini}
\label{main3}
\end{table*}

\section{Experiments} \label{expmt}

In this section, we conduct experiments on two open-source model families: Qwen2.5 \citep{qwen} and Llama3 \citep{llama3}. For each family, we start from their respective base models.

\subsection{Experiment Details}

\vspace{5pt}
\noindent
{\bf Standard SFT.} For each base model, we establish a baseline by performing standard SFT on the question-proof pairs for 3 epochs.

\vspace{5pt}
\noindent
{\bf Progressive Multi-Stage SFT.} We implement two variants of our proposed progressive training strategy as follows:
\begin{itemize}[itemsep=1pt, parsep=2pt, leftmargin=10pt,topsep=2pt]
    \item \textbf{Full Three-Stage Training}: This variant follows the complete curriculum described in Section \ref{sft}. For a given base model, we first fine-tune it on the theorem-proof pairs for 3 epochs, then on the theorem-sketch-proof for another 3 epochs, and finally on the full hierarchical data for a final 3 epochs.
    
    \item \textbf{Two-Stage Training}: This is a streamlined variant where we skip the intermediate theorem-sketch-proof stage. The motivation for this design is that the core part of our hierarchical structure is the insight and the conceptual jump from generating a full proof to generating a proof sketch might be relatively small in terms of reasoning difficulty. Therefore, removing the proof sketch part may not affect training and could also help save some token cost.
    % {\color{red}(The explanation here is not clear.)}
\end{itemize}

\vspace{5pt}
\noindent
{\bf \texttt{InsightPO} Post-Training.} Building on top of the three-stage SFT models, we conduct \texttt{InsightPO} post-training as described in Section~\ref{InsightPO}. We use \texttt{DeepInsightTheorem} as the training set and DeepSeek-R1 as the verifier. The verifier evaluates insight quality, logical validity, completeness, and clarity with weights of 30\%, 30\%, 25\%, and 15\%, respectively. For each dimension, the verifier outputs a score from $\{0, 0.1, 0.2, \ldots, 1.0\}$; rather than using only the argmax score token, we extract the logits over the 11 score values, apply softmax normalization, and use the probability-weighted expected score $\sum_{i=0}^{10} p_i(i/10)$ as the dimension score before computing the weighted total. We train for 400 steps with batch size 128, group size 16, and a maximum rollout length of 8192 tokens. Following prior work on group-relative policy optimization \citep{r1}, we do not apply KL regularization. The full verifier prompt is provided in Appendix~\ref{prompts}, and the test score dynamics during \texttt{InsightPO} training are shown in Figure~\ref{fig:grpo_curves} in Appendix~\ref{add_result}.

\vspace{5pt}
\noindent
{\bf Evaluation Protocol.} 
% To reduce potential bias from the LLM verifier, we additionally use a small stratified human-audit set as a calibration and sanity-check mechanism. Human reviewers inspect the same hierarchy-aware dimensions used by the verifier, including insight quality, logical validity, completeness, and clarity. Details are provided in Appendix~\ref{eva}.
We evaluate generated proofs with an LLM-as-Judge protocol detailed in Appendix~\ref{eva}. The automatic evaluation scores final proofs along logical validity, completeness, and clarity, and aggregates judgments from DeepSeek-R1 and o3-mini to reduce single-judge bias. Since our training and evaluation both rely on LLM-based verification, we further conduct a stratified human audit as a calibration and sanity-check mechanism. Human reviewers inspect sampled outputs across model families, benchmarks, and score ranges, checking whether the proof realizes the intended insight and whether any essential derivation is missing.

\vspace{5pt}
\noindent
Moreover, we compare our method with two groups of baselines. The first group contains several open-source models, many of which have been further enhanced via RL-based post-training; we compare them with our fine-tuned Qwen2.5-7B model. The second group contains three structural mathematical reasoning baselines, including SELF-DISCOVER, Least-to-Most Prompting, and Plan-and-Solve Prompting. These  baselines are all trained on prompt-proof pairs $(q,p)$. More baseline settings are detailed in Appendix~\ref{structured_prompting_baselines}.

% Given the lack of baselines for informal theorem proving, 
% we also evaluate the performance of 7B-level base models  of the two series on three benchmarks to demonstrate the extent of improvement achieved by our method over the baseline.

% \vspace{-0.1cm}

\subsection{Main Results}
{\bf Our framework achieves better performance}. The main results are presented in Table \ref{main3} and Table \ref{main}. Table \ref{main3} shows the average scores from both DeepSeek-R1 and o3-mini, while Table \ref{main} presents the scores from DeepSeek-R1. These results show that after training for insightful hierarchical reasoning, for both variants of our proposed progressive training strategy, the models of different sizes demonstrate superior performance to base models on all three benchmarks. This verifies that, via insightful reasoning, the model's reasoning capability improves. 

% \begin{figure}
%     \centering
%     \begin{subfigure}{\linewidth}
%         \centering
%         \includegraphics[width=0.8\linewidth, height=4cm]{711773717025_.pic_hd.jpg}
%         % \caption{}
%         \label{p1a}
%     \end{subfigure}
    
%     \vspace{0.5em} 
    
%     \begin{subfigure}{\linewidth}
%         \centering
%         \includegraphics[width=0.8\linewidth, height=4cm]{701773717021_.pic_hd.jpg} 
%         % \caption{}
%         \label{p1b}
%     \end{subfigure}
    
%     \caption{Max scores in evaluation}
%     \label{p1}
% \end{figure}

% Since the improvement of our method over the baseline is more significant than that of the baseline over the Base models, as can be seen in Table \ref{main2}, this indicates that the performance gain achieved by our method is meaningful and substantial. 
In addition, Figure \ref{p1} in Appendix \ref{add_result} reports the best scores for each benchmark, which are pushed higher when incorporating insight, especially for smaller backbones such as the 1.5B model. This implies that insight-guided thinking can help break the reasoning ability ceiling for small models.

\vspace{3pt}

\noindent
{\bf \texttt{InsightPO} further boosts performance across all models.}
As shown in the ``\texttt{DeepInsight}'' rows in Table~\ref{main} and Table~\ref{main3}, \texttt{InsightPO} post-training on top of the three-stage SFT models yields consistent and substantial improvements on all benchmarks. For the Qwen2.5-7B model, \texttt{InsightPO} achieves an average score of 34.32, representing a relative improvement of 32.9\% over the three-stage SFT baseline.

\vspace{3pt}

\noindent
{\bf Our framework outperforms  SOTA models.}
The comparison can be seen in Table \ref{main1} in Appendix \ref{E2}. These results demonstrate that, after only  SFT, our model already achieves performance that is competitive with those of SOTA models of the same size and extensive post-training, and notably outperforms them after \texttt{InsightPO} training. For the details of evaluation, please see Appendix \ref{eva}.

\vspace{3pt}

\noindent
{\bf Comparisons with structured prompting baselines.}
The  results further show that generic  decomposition or planning improves over plain SFT but remains consistently below \texttt{DeepInsight}. This suggests that the main gain does not come only from asking the model to produce a plan at inference time, rather, it comes from explicitly learning theorem-specific insight hierarchies during training and then refining them with \texttt{InsightPO} . Detailed settings and results are provided in Appendix~\ref{structured_prompting_baselines}.

\section{Conclusion}
\vspace{-0.1cm}

% This paper studies informal theorem proving and identifies core techniques as a key bottleneck in model training. To address this, 
In this paper, we propose \texttt{DeepInsight}, a complete training framework consisting of: (1) \texttt{DeepInsightTheorem}, a hierarchical dataset with explicitly extracted core techniques; (2) a progressive curriculum-style SFT strategy that trains LLMs for insightful reasoning; and (3) \texttt{InsightPO}, a GRPO-style reinforcement learning algorithm with a probability-weighted LLM verifier and hierarchy-aware rewards. Comprehensive evaluations show consistent and substantial improvements over strong baselines on challenging benchmarks.

% \section*{Limitations}
% While our insight-aware hierarchy improves informal theorem proving, several limitations remain.
% First, our
% $(q, \{t_i\}, s, p)$
%  annotations are produced
%  with LLM-assisted decomposition, which may introduce noise; future work will strengthen this
%  pipeline via self-consistency, cross-model agreement, and selective human auditing. Second, our
%  evaluation relies on LLM-as-Judge, which can
%  be prompt-sensitive; we will incorporate  hybrid automatic checks
%  (e.g., structure/consistency and domain-specific
%  symbolic validation when applicable). Finally, our
%  experiments cover a limited set of benchmarks and
%  informal proofs cannot guarantee correctness; we
%  plan broader out-of-distribution evaluations and
%  tighter integration with verification-oriented feedback (e.g., checking key subclaims or partially formalized steps).


% \section*{Ethical Statement}
% This work enhances LLM capabilities for informal
% theorem proving to support research and education,
%  though it does not guarantee formal correctness.
% Because the model can generate convincing yet potentially erroneous proofs, we recommend expert
%  oversight for high-stakes applications and appropriate safeguards against academic dishonesty. Additionally, as our evaluation relies on LLM judges,
%  results should be interpreted with awareness of potential biases and limitations in detecting subtle
% mathematical flaws.

\bibliography{main}

\clearpage
\appendix

\section{Related Work}
\label{sec:related_work}

\paragraph{Formal theorem proving.}
Formal automated theorem proving (ATP) relies on proof assistants such as Lean, Coq, and Isabelle (\citealp[ ]{zheng2022minif2f} \citealp[ ]{fimo}\citealp[]{putnam}), where correctness is enforced by a strict formal system.
Recent work integrates large language models with formal proof environments to combine natural-language generation with machine-checkable verification (\citet{gloeckle2024abel}, \citet{hu2025minictxneuraltheoremproving}, \citet{lin2025leanstarlearninginterleavethinking}, \citet{poesia2024learningformalmathematicsintrinsic}).
Representative benchmarks include miniF2F for cross-system olympiad-level evaluation \citep{zheng2022minif2f}, as well as more challenging datasets that target IMO-/Putnam-style mathematics \citep{liu2023fimo,putnam}.
While formal approaches offer strong verifiability, they often face (i) a substantial gap between human mathematical exposition and formal languages, and (ii) a large proof search space that benefits from stronger decomposition, planning, and retrieval.

\paragraph{Informal theorem proving.}
Informal theorem proving generates proofs directly in natural language and standard mathematical notation (e.g., \LaTeX), which aligns well with typical LLM pretraining.
NaturalProofs constructs natural-language theorem--proof corpora \citep{napro}, and NaturalProver studies grounded proof generation in reference/retrieval settings \citep{napro}.
Tencent's DeepTheorem further advances IMO-level informal proving with large-scale data and reinforcement-learning-style training recipes \citep{zhang2025deeptheorem}. \citet{lu2025solvinginequalityproofslarge} focuses specifically on inequality proofs.
Despite this progress, many systems remain largely end-to-end and do not explicitly identify or control the core techniques that drive a proof.

\paragraph{Hierarchical Reasoning.} Hierarchical reasoning for LLMs decomposes complex problems into manageable sub-problems, enhancing reasoning accuracy and efficiency in complex scenarios. HyperTree Planning \citep{gui2025hypertreeplanningenhancingllm} constructs a structured, high-level outline before generating details. ReasonFlux \citep{yang2025reasonfluxhierarchicalllmreasoning} goes beyond raw text generation by scaling "thought templates". Plan-and-Solve Prompting \citep{wang2023planandsolvepromptingimprovingzeroshot} first asks the model to form an explicit plan before producing the final answer, while Least-to-Most Prompting \citep{zhou2023leasttomostpromptingenablescomplex} decomposes a difficult problem into simpler subproblems and solves them sequentially. SELF-DISCOVER \citep{zhou2024selfdiscoverlargelanguagemodels} enables LLMs to compose reasoning structures from atomic modules. CoGer \citep{hu2025fastslowcognitiveinspiredelastic} draws on Bloom's taxonomy to classify query complexity into four levels and trains an agent via reinforcement learning to dynamically select appropriate reasoning strategies.
Another line of work lies in modifying model architectures to support hierarchical processing such as
\citet{wang2025hierarchicalreasoningmodel}, \citet{sun2025hierarchicalmemoryhighefficiencylongterm}.

In addition, several works focus on hierarchical reasoning for theorem proving. For example, \citet{dong2024formaltheoremprovingrewarding} proposes a reinforcement learning-based training algorithm that incentivizes LLMs to hierarchically decompose theorems into lemmas. \citet{ye2024reasoning} decomposes complex theorem-proving tasks into small, achievable subgoals to abstract formal proof steps.

The key idea in these works is to decompose a complex problem into smaller parts, which corresponds to the proof-sketch component (Section \ref{4.1}) in our work. Our hierarchical method instead identifies core techniques, operating at a higher level than a sketch. The work most similar to ours is Reason-Flux \citep{yang2025reasonfluxhierarchicalllmreasoning}. However, it depends on constructing and retrieving from a template library for theorem proving, which does not endow the LLM itself with intrinsic hierarchical thinking capabilities. In our work, the LLM has been trained to develop insightful thinking, which essentially empowers the LLM with high-level thinking patterns.

% \section{Proof Details}
% \label{proof}

\section{Proof Details Omitted in Section \ref{3.1} }
\label{a1}
Recall that $\pi_{\text{TP}}(p|q)$ is the probability of $\pi_{\text{TP}}$ generating a valid proof $p$ for question $q$. The general form of a proof is shown in \eqref{p}. We let $k$ be the number of core techniques in a proof $p$.
Since we assume $\pi_{\text{TP}}(t_i |u_i,  q) < \delta$ for all $i\leq k$, by Bayes' rule, we have
\begin{align}
&  \pi_{\text{TP}}(p|q)  \nonumber\\
& = \prod_{i\leq k}  \left [ \pi_{\text{TP}}(r_i |t_{i-1},u_{i-2}, q) \pi_{\text{TP}}(t_i |u_i,q) \right ] \nonumber \\
& \leq \prod_{i\leq k}  \delta \cdot \pi_{\text{TP}}(r_i |t_{i-1},u_{i-2}, q) \label{a17} \\
& \le  \delta^k,  \label{a18}
\end{align}
where the prefix $u_i$ is defined by $u_i := (r_1,t_1,\cdots,r_i)$. Note that \eqref{a17} is obtained by the assumption $\pi_{\text{TP}}(t_i |u_i,  q) < \delta$ for all $i\leq k$ and \eqref{a18} is due to $\pi_{\text{TP}}(r_i |t_{i-1},u_{i-2}, q)\leq 1$.

\begin{figure*}
 \begin{evaluationbox}{O3-mini's Evaluation-DeepSeek-R1}
    % 第一部分：简短列表
    \begin{enumerate}[label=\arabic*., leftmargin=*, nosep]
        \item \textbf{mixed}
        \item \textbf{mixed}
        \item \textbf{incomplete}
    \end{enumerate}

    \vspace{1em}
    Below is an explanation for each answer:

    % 第二部分：详细解释列表

    \begin{enumerate}[label=\arabic*., leftmargin=*, itemsep=1em]
        \item \textbf{Mixed:} \\
        The insight mentions several techniques---substitution (setting $y=1$ or $x=1$), exploring injectivity/surjectivity, and even hinting at Cauchy-type ideas as a route to establishing linearity. However, these ideas remain quite standard for this type of functional equation problem. They are useful, but the answer doesn't pinpoint any single, non-trivial transformation or method that clearly stands out as a breakthrough insight; it rather relies on a collection of standard tactics. Thus, while not entirely shallow, the ideas include both non-shallow observations and standard, almost routine deductions.

        \item \textbf{Mixed:} \\
        The insight does provide some accurate general directions, which are key in many functional equation proofs. However, the description remains somewhat vague and does not detail enough concrete, non-trivial steps. This mix of accurate generalizations and lack of detailed construction leads to a "mixed" evaluation.

        \item \textbf{Incomplete:} \\
        While the insight mentions common techniques, it does not capture all potential core ideas needed to solve the problem. There could be additional subtleties---such as a rigorous step-by-step approach to prove that $f$ must be affine or specific manipulations that pivot the equation to reveal the structure $f(x) = x+1$. Therefore, the core techniques mentioned are not fully comprehensive to cover all aspects of the solution; they provide an initial guide but remain incomplete.
    \end{enumerate}
\end{evaluationbox}
 \caption{o3-mini's evaluation response for the insight generated by DeepSeek-R1}
\label{o3r1}
\end{figure*}

\begin{figure*}
\begin{evaluationbox}{O3-mini's Evaluation-Gemini2.5-Flash}
    % 第一部分：简短列表
    \begin{enumerate}[label=\arabic*., leftmargin=1.5em, nosep]
        \item \textbf{mixed}
        \item \textbf{simple scratch}
        \item \textbf{incomplete}
    \end{enumerate}

    \vspace{0.8em}
    Below is the explanation for each:

    % 第二部分：详细解释
    \begin{enumerate}[label=\arabic*., leftmargin=1.5em, itemsep=0.8em, topsep=0.5em]
        \item \textbf{Mixed:} \\
        The insight mentions exploring properties like injectivity, surjectivity, particular substitutions (e.g., \(x=1\), \(y=1\)), and fixed points, and notes that the structure of the equation suggests multiplicative features. While these are useful starting points, they represent standard techniques in functional equation problems. They do hint at a deeper structure but do not provide a concrete key step that addresses the heart of obtaining \(f(x)=x+1\). Thus, there exists a mix of non-trivial ideas combined with basic, common observations.

        \item \textbf{Simple Scratch:} \\
        The insight lists several exploratory steps but does so in a rather vague manner without a precise roadmap tailored to this particular question. The ideas are on the right track but lack the detail needed for a detailed, step-by-step core technique application.

        \item \textbf{Incomplete:} \\
        While the insight addresses several common techniques for approaching functional equations, it does not cover all the essential steps needed to rigorously derive \(f(x)=x+1\). Additional deeper reasoning, such as deducing specific values (like evaluating \(f(1)\)) and showing uniqueness or employing clever substitutions that eventually simplify the functional relationship, is necessary. Thus, the list of techniques provided is not comprehensive enough to fully solve the problem without filling in significant logical and algebraic deductions.
    \end{enumerate}
\end{evaluationbox}
\caption{o3-mini's evaluation response for the insight generated by Gemini 2.5 Flash}
\label{o3ge}
\end{figure*}

\begin{figure*}
  % \label{fig:experiments}
\begin{mathanalysisbox}
    \textbf{Question:} \\
    Let $X$ be a non-negative random variable and $g : [0, \infty) \to [0, \infty)$ be a monotone nondecreasing function such that $g(x) \to \infty$ as $x \to \infty$. If $g(X)$ is integrable, i.e., $E[g(X)] < \infty$, prove that $\lim_{x \to \infty} g(x)P(X > x) = 0$.

    \medskip
    \textbf{Response:} \\
    \textless tech\textgreater \\
    Let's analyze the conditions in this question. The monotonicity and divergence of $g$ naturally suggest linking the tail event $\{X > x\}$ to the growth of $g(X)$, forming a crucial inclusion that leverages $g$'s behavior. The integrability of $g(X)$ prompts Markov's inequality to bound the tail probability by an expectation, exploiting the non-negativity and finite mean. The pointwise convergence to zero, combined with the integrable bound $g(X)$, calls for the Dominated Convergence Theorem to justify limit-interchange, which is indispensable since the expectation structure captures the asymptotic decay. Therefore, the potential techniques are summarized as: \\
    \textless construction\textgreater: Event inclusion $\{X > x\} \subseteq \{g(X) \ge g(x)\}$ via monotonicity of $g$ \\
    \textless theorem call\textgreater: Markov's inequality and Dominated Convergence Theorem \\
    \textless transformation\textgreater: no \\
    \textless /tech\textgreater

    \medskip
    \textless sketch\textgreater
    \begin{enumerate}[label=\arabic*., nosep, leftmargin=2em]
        \item Use the monotonicity and divergence of $g$ to establish $\{X > x\} \subseteq \{g(X) \ge g(x)\}$, implying $P(X > x) \le P(g(X) \ge g(x))$.
        \item Apply Markov's inequality to the nonnegative integrable $g(X)$, yielding $g(x)P(X > x) \le E[g(X)\mathbf{1}_{\{g(X) \ge g(x)\}}]$.
        \item Observe pointwise convergence $g(X)\mathbf{1}_{\{g(X) \ge g(x)\}} \to 0$ due to $g(x) \to \infty$, and dominate by $g(X)$.
        \item Invoke the Dominated Convergence Theorem to show $E[g(X)\mathbf{1}_{\{g(X) \ge g(x)\}}] \to 0$.
        \item Conclude $\lim_{x \to \infty} g(x)P(X > x) = 0$ via the squeeze theorem on $0 \le g(x)P(X > x) \le E[g(X)\mathbf{1}_{\{g(X) \ge g(x)\}}]$.
    \end{enumerate}
    \textless /sketch\textgreater

    \medskip
    \textless proof\textgreater \\
    Since $g$ is monotone nondecreasing, for any $x \ge 0$, the condition $X > x$ implies $g(X) \ge g(x)$. This gives the event inclusion:
    \[ \{X > x\} \subseteq \{g(X) \ge g(x)\}, \]
    which, by monotonicity of probability, yields:
    \[ P(X > x) \le P(g(X) \ge g(x)). \]
    As $g(X)$ is nonnegative and $E[g(X)] < \infty$, Markov's inequality for $a = g(x) > 0$ (for large $x$ since $g(x) \to \infty$) states:
    \[ P(g(X) \ge g(x)) \le \frac{E[g(X)\mathbf{1}_{\{g(X) \ge g(x)\}}]}{g(x)}. \]
    Multiplying by $g(x)$ and using the earlier inequality, we obtain:
    \[ g(x)P(X > x) \le g(x)P(g(X) \ge g(x)) \le E[g(X)\mathbf{1}_{\{g(X) \ge g(x)\}}]. \]
    Fix $\omega \in \Omega$. As $x \to \infty$, $g(x) \to \infty$ so $g(x) > g(X(\omega))$ for large enough $x$, forcing $\mathbf{1}_{\{g(X(\omega)) \ge g(x)\}} = 0$. Thus:
    \[ \lim_{x \to \infty} g(X(\omega))\mathbf{1}_{\{g(X(\omega)) \ge g(x)\}} = 0, \quad \forall \omega. \]
    Moreover, $0 \le g(X)\mathbf{1}_{\{g(X) \ge g(x)\}} \le g(X)$, and $E[g(X)] < \infty$ by hypothesis. By the Dominated Convergence Theorem:
    \[ \lim_{x \to \infty} E[g(X)\mathbf{1}_{\{g(X) \ge g(x)\}}] = E\left[ \lim_{x \to \infty} g(X)\mathbf{1}_{\{g(X) \ge g(x)\}} \right] = E[0] = 0. \]
    Finally, sandwiching with:
    \[ 0 \le g(x)P(X > x) \le E[g(X)\mathbf{1}_{\{g(X) \ge g(x)\}}] \xrightarrow{x \to \infty} 0, \]
    we conclude $\lim_{x \to \infty} g(x)P(X > x) = 0$ as required. \\
    \textless /proof\textgreater
\end{mathanalysisbox}
  \caption{A data example from \texttt{DeepInsightTheorem} }
  \label{figdata}
\end{figure*}

\newpage
\section{Additional Experimental Results}
\label{add_result}
\subsection{Experiments in Section \ref{2.3}} 
\label{exp2.3}
We conduct an experiment on how insightful the recent advanced commercial LLMs can be. We choose Gemini 2.5 Flash and DeepSeek-R1 to generate insights of several problems. The problems are chosen from four categories including algebra, analysis, combinatorics and number theory, each contains 10 mathematical competition level questions from Putnam and FIMO dataset. The insight generation prompt is listed in Appendix \ref{prompts} titled 'Insightfulness Evaluation'. Then we evaluate these generated insight by o3-mini through the prompt titled 'Insight Evaluation'. The results show that most of the insights are basically listing the general standard techniques according to the conditions in questions and can not grasp the essential approaches. See  a randomly chosen  evaluation result of the following question. The evaluations for both models are shown in Figure \ref{o3r1} and Figure \ref{o3ge}.

\vspace{5pt}
\noindent
{\bf Question}: Let $\mathbb{R}^{+}$be the set of positive real numbers. Determine all functions $f: \mathbb{R}^{+} \rightarrow \mathbb{R}^{+}$ such that, for all positive real numbers $x$ and $y$,
$$f(x+f(x y))+y=f(x) f(y)+1, $$
the final answer is $f(x)=x+1$.

\vspace{5pt}
\noindent{\bf Insights generated by DeepSeek-R1}: The functional equation involves terms like \(f(x + f(xy))\) and \(f(x)f(y) + 1\), suggesting the use of **substitution techniques** such as setting \(y = 1\) or \(x = 1\) to simplify and obtain relations involving \(f(1)\).  
A potential approach is to **explore injectivity/surjectivity** by fixing one variable and comparing expressions, then using **Cauchy-type ideas** after establishing \(f\) is linear or affine.  
The structure \(f(x + f(xy)) - f(x)f(y) = 1 - y\) hints at eventually proving \(f\) is of the form \(ax + b\) and then determining constants via substitution into the original equation.

% \newpage
\vspace{5pt}
\noindent
{\bf Insights generated by Gemini 2.5 Flash}: The presence of $f(x+f(xy))$ and $f(x)f(y)$ suggests looking for properties like injectivity, surjectivity, or particular values of $f$. Substituting specific values (e.g., $x=1$, $y=1$) or analyzing the fixed points of $f$ might be crucial. The term $f(xy)$ combined with $f(x)f(y)$ often points to multiplicative properties or comparison with Cauchy's functional equation variants.

\vspace{5pt}

We see that most of the insights generated by both Gemini 2.5 Flash and DeepSeek-R1 for this question are merely generic analysis, without getting to the essence. The insights generated by Gemini are even less accurate, and can even be regarded as ``simple scratch.''

\subsection{Additional Baseline Settings}
\label{structured_prompting_baselines}
We evaluate three additional baselines under the same theorem-proving setting as our main experiments. The full prompts for these baselines are provided in Appendix~\ref{prompts}, under ``Plan-and-Solve Baseline Prompt'', ``Least-to-Most Prompt'', and ``SELF-DISCOVER Prompt''. For Plan-and-Solve, the model is prompted to first write a concise proof plan and then generate the final proof conditioned on that plan. For Least-to-Most Prompting, the model is prompted to decompose the theorem into a sequence of easier subclaims, solve them in order, and combine the intermediate results into the final proof. For SELF-DISCOVER, the model is prompted to select and compose general reasoning modules into a task-specific reasoning structure before generating the proof.

For fairness, all three baselines start from the same standard SFT checkpoint trained only on prompt-proof pairs $(q,p)$. They are applied only at test time on Qwen2.5-7B and Llama3-8B. All generations use the same benchmark split, decoding budget, proof extraction rule, LLM-as-judge protocol, and human-audit calibration as our main experiments. Table~\ref{structured_prompt_table} reports the resulting comparison.

\begin{table*}[t]
\centering
\renewcommand{\arraystretch}{0.85}
\setlength{\tabcolsep}{8pt}
\begin{tabular}{l l c c c c}
\toprule
\textbf{Models} & \textbf{Methods} & \textbf{FIMO} & \textbf{Putnam} & \textbf{HMMT} & \textbf{Avg.} \\
\midrule
\multirow{4}{*}{Qwen2.5-7B}
& SFT$(q,p)$ + Plan-and-Solve & 22.40 & 48.70 & 20.90 & 30.67 \\
& SFT$(q,p)$ + Least-to-Most & 23.65 & 47.90 & 17.35 & 29.63 \\
& SFT$(q,p)$ + SELF-DISCOVER & 22.95 & 51.60 & 16.70 & 30.42 \\
& \texttt{DeepInsight}& \textbf{27.80} & \textbf{55.50} & \textbf{24.60} & \textbf{35.97} \\
\midrule
\multirow{4}{*}{Llama3-8B}
& SFT$(q,p)$ + Plan-and-Solve & 17.35 & 41.80 & 12.60 & 23.92 \\
& SFT$(q,p)$ + Least-to-Most & 16.80 & 44.20 & 13.75 & 24.92 \\
& SFT$(q,p)$ + SELF-DISCOVER & 19.45 & 43.60 & 14.95 & 26.00 \\
& \texttt{DeepInsight}& \textbf{22.10} & \textbf{50.20} & \textbf{20.30} & \textbf{30.87} \\
\bottomrule
\end{tabular}
\caption{Comparison with additional baselines on 7B/8B models. All baselines use the standard SFT checkpoint trained only on $(q,p)$ and are applied at test time.}
\label{structured_prompt_table}
\end{table*}

\subsection{Parameter Settings in Section \ref{expmt}} \label{E2}
Here we provide main parameters setting for our experiments. For both model series, we use the learning rate of 2e-5, and training batch size 256. The maximal sequence length is set to be 4096 for all experiments on the base dataset, and 8192 for all on \texttt{DeepInsightTheorem}. We also conducted additional experiments with the baseline's max sequence length set to 8192 for completion. See Table \ref{base8092}
for details.
\begin{table}[htbp]
\centering
\setlength{\tabcolsep}{4.6pt}
\begin{tabular}{l c c c c}
\toprule
Methods & FIMO & Putnam & HMMT & Avg. \\
\midrule
Qwen2.5-7B & 14.98 & 36.05 & 15.24 & 22.09 \\
Llama3-8B  & 13.29 & 38.84 & 11.99 & 21.37 \\
\bottomrule
\end{tabular}
\caption{Evaluation of baseline with max length 8192}
\label{base8092}
\end{table}

\begin{table*}[t]
  \centering
  \renewcommand{\arraystretch}{0.8} % 增加行高
  \setlength{\tabcolsep}{15pt} % 增加列间距
  
  \begin{tabular}{l l c c c c}
    \toprule
    \textbf{Models}  & \textbf{FIMO} & \textbf{Putnam} & \textbf{HMMT} & \textbf{Avg.} \\
\midrule
    
    Qwen2.5-Inst-7B
      & 15.29 & 42.39 & 17.86 & 25.12 \\
    \addlinespace[2pt]

    \midrule
    Qwen2.5-Math-Inst-7B
       & 17.01 & 41.06 & 16.57 & 24.92 \\
    \addlinespace[2pt]

    \midrule
    DS-Prover-v1.5-RL-7B
      & 17.39 & 42.00 & 13.68 & 24.36 \\
    \addlinespace[2pt]

    \midrule
    DS-Prover-v2-7B
      & 16.25 & 41.50 & 15.43 & 24.39 \\
    \addlinespace[2pt]

    \midrule
    \midrule
    Ours (2-stage)
      & 16.33 & 43.34 & 15.59 & 25.09 \\
    \addlinespace[2pt]

    \midrule
    Ours (3-stage)
       & 18.03 & 41.67 & 17.78 & 25.83 \\
    \addlinespace[2pt]

    \midrule
    Ours (\texttt{DeepInsight})
       & {\bf 26.10} & {\bf 53.70} & {\bf 23.15} & {\bf 34.32} \\

    \bottomrule
  \end{tabular}
  
  \vspace{6pt}
  \caption{
    Evaluation score by Deepseek-r1 comparison against Open-Source Baselines
    % {\color{red}(You did not highlight all best results.)}
  }
  \label{main1}
\end{table*}

\begin{figure}
    \centering
    \begin{subfigure}{\linewidth}
        \centering
        \includegraphics[width=0.80\linewidth, height=4.2cm]{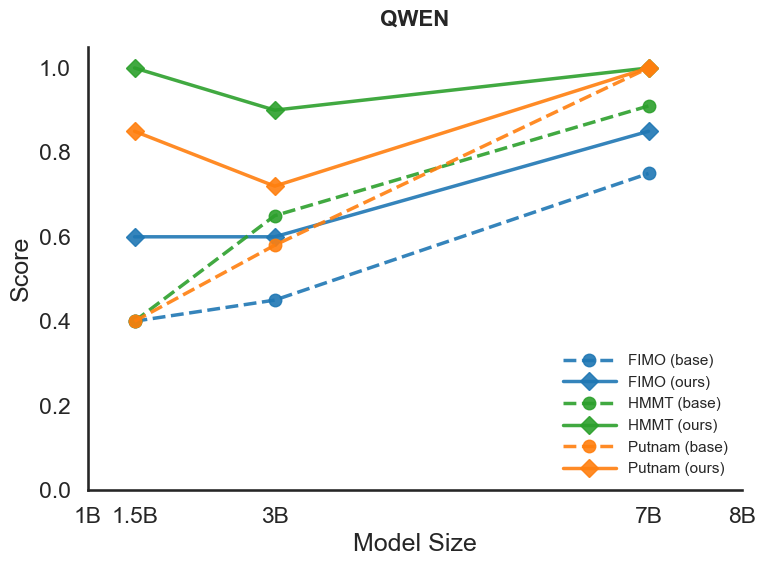}
        % \caption{}
        \label{p1a}
    \end{subfigure}
    
    \vspace{0.5em} 
    
    \begin{subfigure}{\linewidth}
        \centering
        \includegraphics[width=0.80\linewidth, height=4.2cm]{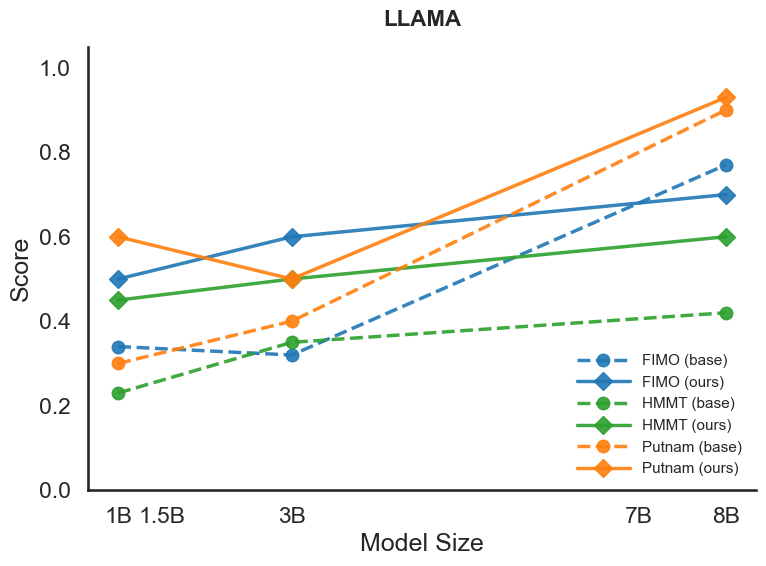} 
        % \caption{}
        \label{p1b}
    \end{subfigure}
    
    \caption{Max scores in evaluation across benchmarks.}
    \label{p1}
\end{figure}

\begin{figure*}[t]
    \centering
    \includegraphics[width=\textwidth]{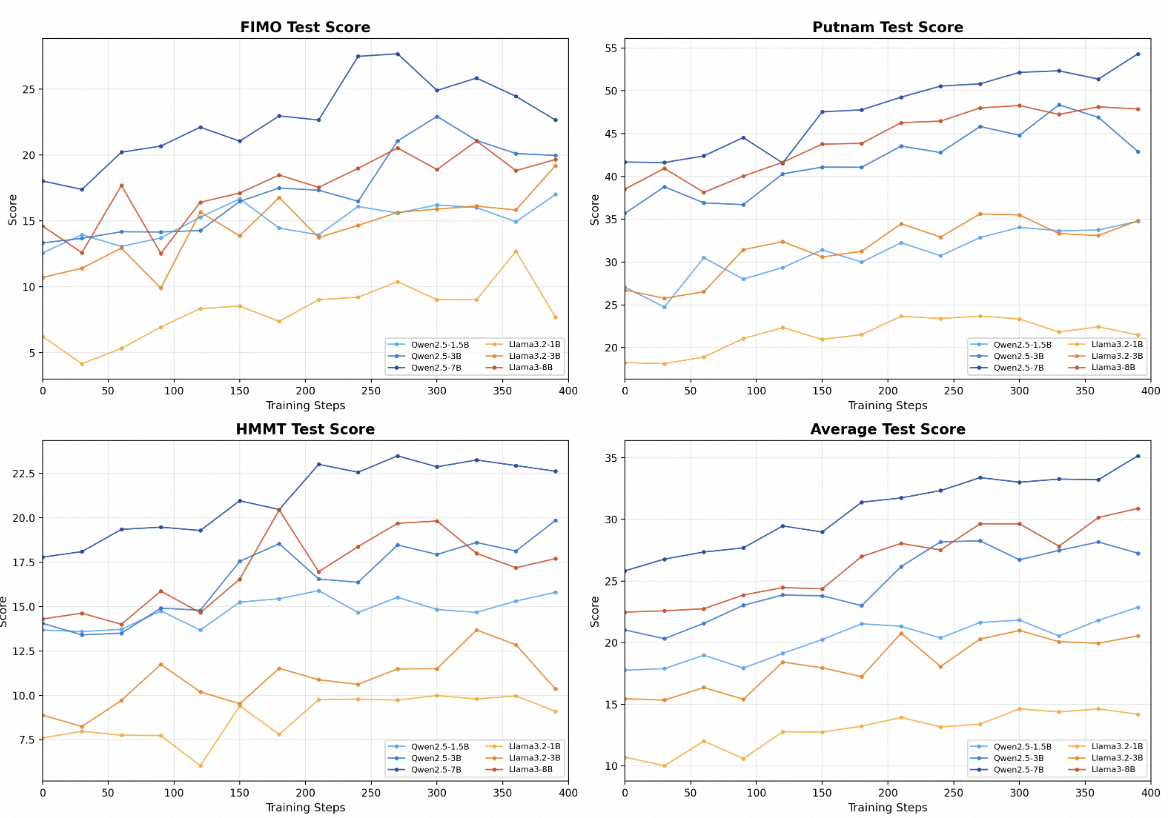}
    \caption{Test score curves during \texttt{InsightPO} training. }
    \label{fig:grpo_curves}
\end{figure*}

\begin{table*}[t]
  \centering
  \renewcommand{\arraystretch}{0.4} % 增加行高
  \setlength{\tabcolsep}{12pt} % 增加列间距
  \begin{tabular}{@{}l l c c c c}
    \toprule
    \textbf{Models} & \textbf{Methods} & \textbf{FIMO} & \textbf{Putnam} & \textbf{HMMT} & \textbf{Avg.} \\
\midrule
    
    \multirow{4}{*}{Qwen2.5-1.5B}
      & Base & 11.17 & 25.39 & 11.25 & 15.94 \\
    \cmidrule(lr){2-2}
      & Two-stage & 11.81 & 28.72 & 11.80 & 17.44 \\
      & Three-stage & 12.57 & 27.06 & 13.68 & 17.77 \\
    \cmidrule(lr){2-2}
      & \texttt{DeepInsight}& {\bf 16.10} & {\bf 33.50} & {\bf 15.90} & {\bf 21.83} \\
    \addlinespace[2pt]

    \midrule
    \multirow{4}{*}{Qwen2.5-3B}
      & Base & 11.92 & 31.87 & 12.69 & 18.83 \\
    \cmidrule(lr){2-2}
      & Two-stage & 12.85 & 34.86 & 11.98 & 19.90 \\
      & Three-stage & 13.34 & 35.69 & 14.06 & 21.03 \\
    \cmidrule(lr){2-2}
      & \texttt{DeepInsight}& {\bf 19.50} & {\bf 45.60} & {\bf 18.40} & {\bf 27.83} \\
    \addlinespace[2pt]

    \midrule
    \multirow{4}{*}{Qwen2.5-7B}
      & Base & 15.27 & 36.75 & 14.82 & 22.28 \\
    \cmidrule(lr){2-2}
      & Two-stage & 16.33 & 43.34 & 15.59 & 25.09 \\
      & Three-stage & 18.03 & 41.67 & 17.78 & 25.83 \\
    \cmidrule(lr){2-2}
      & \texttt{DeepInsight}& {\bf 26.10} & {\bf 53.70} & {\bf 23.15} & {\bf 34.32} \\
    \addlinespace[2pt]

    \midrule
    \multirow{4}{*}{Llama3.2-1B}
      & Base & 5.04 & 14.61 & 4.48 & 8.04 \\
    \cmidrule(lr){2-2}
      & Two-stage & 4.19 & 18.26 & 5.78 & 9.41 \\
      & Three-stage & 6.25 & 18.26 & 7.60 & 10.70 \\
    \cmidrule(lr){2-2}
      & \texttt{DeepInsight}& {\bf 9.70} & {\bf 23.30} & {\bf 9.50} & {\bf 14.17} \\
    \addlinespace[2pt]

    \midrule
    \multirow{4}{*}{Llama3.2-3B}
      & Base & 9.29 & 23.90 & 7.04 & 13.41 \\
    \cmidrule(lr){2-2}
      & Two-stage & 11.08 & 24.73 & 9.27 & 15.03 \\
      & Three-stage & 10.72 & 26.73 & 8.89 & 15.45 \\
    \cmidrule(lr){2-2}
      & \texttt{DeepInsight}& {\bf 16.15} & {\bf 34.80} & {\bf 12.25} & {\bf 21.07} \\
    \addlinespace[2pt]

    \midrule
    \multirow{4}{*}{Llama3-8B}
      & Base & 13.48 & 37.68 & 12.25 & 21.14 \\
    \cmidrule(lr){2-2}
      & Two-stage & 14.70 & 40.51 & 14.15 & 23.12 \\
      & Three-stage & 14.62 & 38.51 & 14.30 & 22.48 \\
    \cmidrule(lr){2-2}
      & \texttt{DeepInsight}& {\bf 20.60} & {\bf 48.50} & {\bf 18.70} & {\bf 29.27} \\
    \bottomrule
  \end{tabular}
  
  % \vspace{3pt}
  \caption{
    Evaluation of models trained with our methods and the baseline by DeepSeek-R1
  }
  \label{main}
   % \vspace{-0.3cm}
\end{table*}

\section{Evaluation} \label{eva}
Since our task involves generating informal mathematical proofs by natural language,  we adopt an LLM-as-Judge evaluation protocol. Following the established practice in DeepTheorem \citep{zhang2025deeptheorem}, our evaluation is conducted on a set of challenging benchmarks to test the model's reasoning capability. Specifically, we use theorem-proving problems drawn from FIMO \citep{fimo}, PutnamBench \citep{putnam}, and a newly constructed theorem-proving subset of the Harvard-MIT Mathematics Tournament (HMMT) \citep{hmmt}.

To assess the quality of the generated proofs, we apply the evaluation framework in DeepTheorem but omit a separate ``correctness'' judgment, which is a  fabricated verifiable answer constructed by DeepTheorem that is yet unrelated to the quality of proof itself and hence not applicable to our task. Our evaluation centers on the following three core dimensions of the proof text itself \citep{zhang2025deeptheorem}: 
\begin{itemize}[itemsep=1pt, parsep=2pt, leftmargin=10pt]
    \item {\bf Logical Validity.} Check whether each step follows logically from the preceding step and indicate any logical errors.
    \item {\bf Completeness.} Verify whether all necessary steps are included to fully prove the theorem.
    \item {\bf Clarity.} Assess whether the proof is clear, unambiguous, and well-explained.
\end{itemize}

We use DeepSeek-R1 \citep{r1} as the judge model. For each generated proof, the judge is prompted to analyze and score it on a continuous scale from 0 to 1 for each of the three dimensions above. The final score for a proof is calculated as a weighted average of these three-dimensional scores. The specific weighting scheme and the full prompts used for this evaluation are provided in Appendix \ref{prompts}. 

Using only one LLM as a judge may introduce bias in evaluation. To address this, we also incorporate o3-mini as another LLM judge and combine the scores from both judges to evaluate the performance of our method.
See Table \ref{main3} in  Appendix \ref{E2}
for details.

\vspace{5pt}
\noindent{\bf Human Audit and Calibration.} In addition to multi-judge aggregation, we use a lightweight human-audit protocol to reduce systematic bias in LLM-as-Judge evaluation. We stratify generated proofs by model family, benchmark, and LLM-judge score range, and then sample a small subset for expert inspection. Human reviewers evaluate the same dimensions used by the verifier: insight quality, logical validity, completeness, and clarity. For insight quality, reviewers check whether the generated core techniques capture the key mathematical idea rather than merely restating surface conditions. For proof quality, reviewers check whether the proof sketch and final proof correctly realize the stated techniques and whether any essential derivation is missing. The audited subset is used to calibrate verifier behavior by comparing human scores with LLM scores, checking score deviations across dimensions, and identifying cases where the verifier systematically over-rewards verbose or superficially plausible proofs. As shown in Figure~\ref{fig:human_audit_calibration}, the statistics are computed on a single stratified audit pool aggregated across FIMO, PutnamBench, and HMMT

\begin{figure*}[t]
\centering
\small
\setlength{\tabcolsep}{5pt}
\renewcommand{\arraystretch}{1.05}
\begin{tabular}{l c c c l}
\toprule
\textbf{Score range} & \textbf{Samples} & \textbf{LLM verifier} & \textbf{Human audit} & \textbf{Difference pattern} \\
\midrule
0.0--0.2 & 22 & 0.13 & 0.15 & \rule{0.35cm}{5pt} slight under-scoring \\
0.2--0.4 & 34 & 0.32 & 0.30 & \rule{0.40cm}{5pt} close agreement \\
0.4--0.6 & 41 & 0.51 & 0.46 & \rule{0.85cm}{5pt} gap on incomplete sketches \\
0.6--0.8 & 36 & 0.68 & 0.64 & \rule{0.70cm}{5pt} moderate over-scoring \\
0.8--1.0 & 27 & 0.84 & 0.79 & \rule{0.85cm}{5pt} gap on subtle errors \\
\bottomrule
\end{tabular}
\caption{Human-audit calibration of the LLM verifier, aggregated over the stratified audit subset sampled from FIMO, PutnamBench, and HMMT. Average human scores broadly follow verifier scores across score ranges.}
\label{fig:human_audit_calibration}
\end{figure*}

 % rather than on one benchmark alone. Human scores are broadly consistent with the LLM verifier, especially for clearly incorrect or clearly reasonable proofs. The discrepancies are not simply increasing with the verifier score: mid-score proofs often expose missing proof-sketch steps or unsupported lemmas, while some high-score proofs receive lower human scores because they are fluent but miss subtle cases or contain plausible but non-essential technique descriptions. For such high-disagreement samples, we apply a calibration pass: examples with an absolute human--verifier gap above 0.15 are rechecked by another reviewer, their dimension-level error type is recorded, and the verifier score is corrected by subtracting the average bin-level over-score for the affected dimension. This correction is used only for reporting calibrated audit statistics and for diagnosing verifier bias; the main benchmark tables still follow the same automatic evaluation protocol for all methods.

Finally, since the model trained by our method would generate a formatted output containing three components, we extract only the final proof component for evaluation to ensure a fair comparison with baseline methods.

\section{Data Construction}

As we discussed before, the construction of our data is  based on the DeepTheorem \citep{deeptheorem} by annotating each piece of data with the assistance of DeepSeek-R1 to help generate the core techniques and corresponding proof sketches. After the generation process, we apply a filtering process to remove some data that failed to be annotated due to some reasons like temporary API calling failure or annotated with an undesired structure.   Also we notice some repeated questions in the base dataset hence we also remove those redundant ones. 

\begin{figure*}[t]
    \centering

    \includegraphics[width=\linewidth]{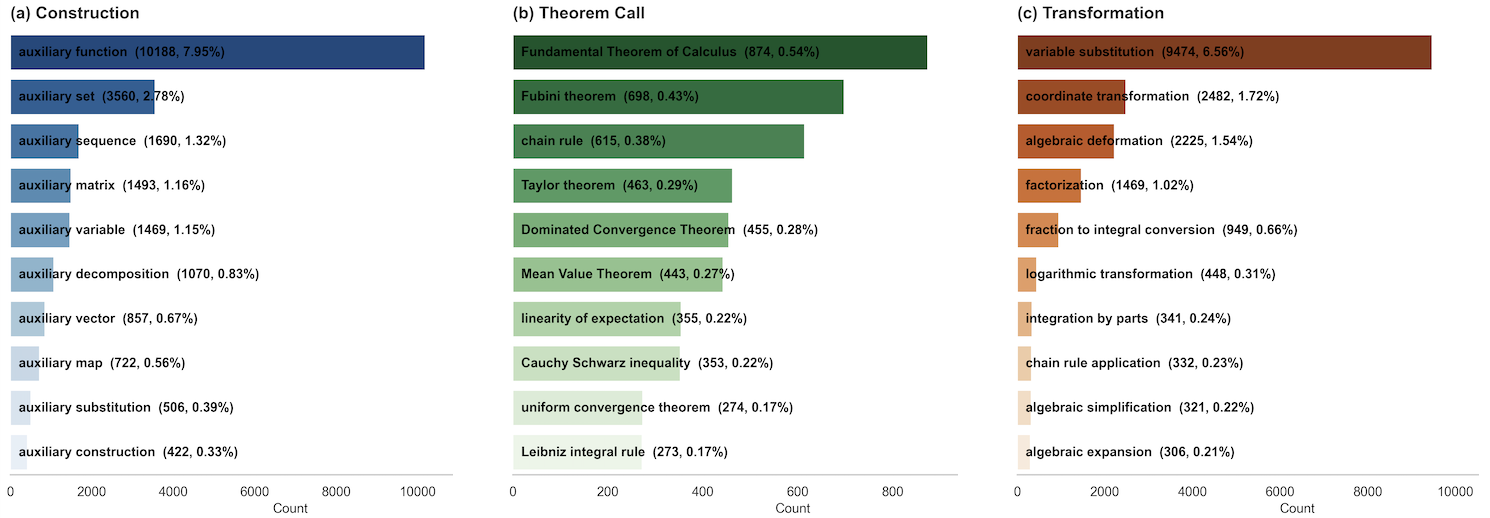}
        % \vspace{0.5em}
    % \includegraphics[width=0.303\linewidth]{bar_2.png}

    \caption{Top technique distribution for each class in \texttt{DeepInsightTheorem}}
    \label{bar2}
\end{figure*}

Finally, we collect 104,751 pieces of hierarchical data with rigorous structure as shown in Figure \ref{figdata}. The distribution of each class of techniques is summarized in Figure \ref{bar2}.

\vspace{5pt}
\noindent{\bf Maximizing Data Utility}. A fundamental principle of \texttt{DeepInsightTheorem} is its self-contained nature. The creation of the hierarchical structure $(q, \{t_i\}, s,  p)$ is achieved solely through the analysis of the information already embedded within the original proofs.  By doing so, we increase the informational density of each training example, offering a powerful pathway to improve data efficiency when scaling high-quality mathematical proof data.
\section{Benchmark Statistics}

Table \ref{bench} is a simple statistic on the number of questions in three benchmarks we use for evaluation. Note that in \citet{deeptheorem}, they manually expanded each benchmark with variants of each question. Here we do not need such a design.

\begin{table}
    \centering
    \setlength{\tabcolsep}{32pt} 
    \renewcommand{\arraystretch}{1.1} % 可选：行距稍微舒服一点
    \begin{tabular}{lc}
      \toprule
      \textbf{Benchmark} & \textbf{Num.}   \\
      \midrule
      FIMO       & 71  \\
      Putnam     & 166 \\ 
      HMMT       & 76  \\
      \bottomrule
    \end{tabular}
    \caption{The number of questions in each benchmark}
    \label{bench}
\end{table}

%     \begin{figure*}[t]
%     \centering
%   \includegraphics[width=\linewidth]{bar_1.png}
%   % \caption{}
%   % \label{}
%   % \label{bar1}
% \end{figure*} 

%     \begin{figure*}
%     \centering
%   \includegraphics[width=0.6\linewidth, height=7cm]{bar_2.png}
%   \caption{}
%   % \label{}
%   \label{bar2}
% \end{figure*} 
\begin{figure*}[t]
    \centering
    \includegraphics[width=\textwidth]{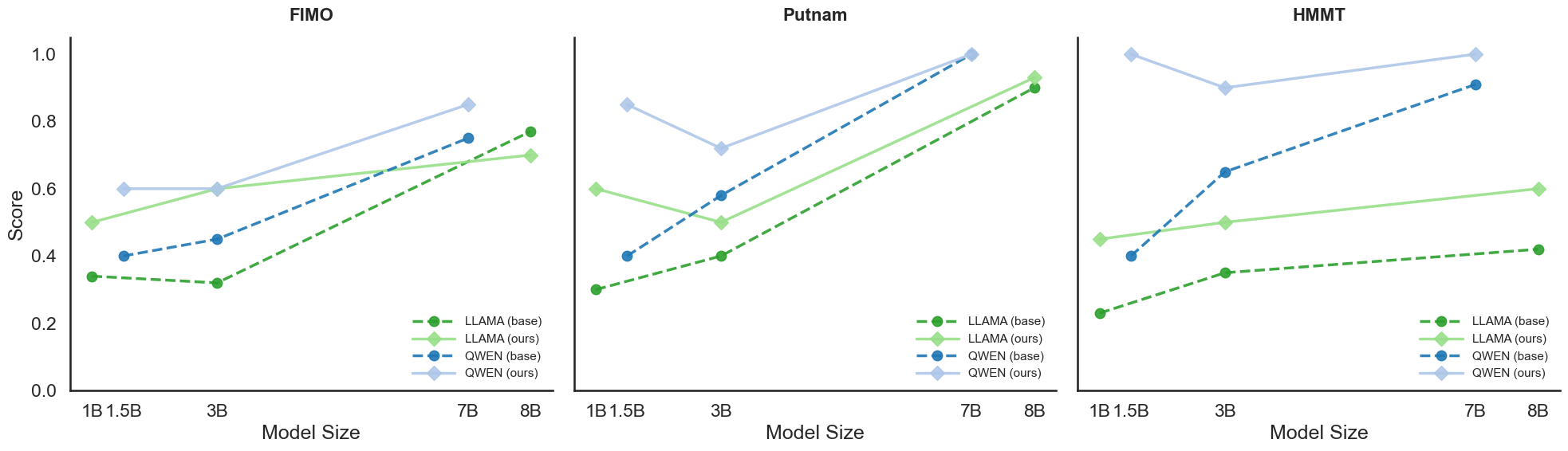}
    \caption{Max scores in evaluation across benchmarks.}
    \label{fig:benchmark_lineplot}
\end{figure*}

\newpage

\section{Qualitative Case Study}

We present a qualitative case study to demonstrate the improved evaluation results. It compares the output generated by the Qwen2.5-7B model trained on \texttt{DeepInsightTheorem} with the baseline output for a sample problem from HMMT.  The chosen problem is ``2025-02-combinatorics-04''.

\vspace{5pt}

\noindent \textbf{Evaluation of Our Method.} We achieved a score of 0.9 for this problem with a minor completeness issue.

% `` \texttt{The proof is exceptionally clear and well-structured. Steps are labeled, transitions are unambiguously defined with precise conditions, and the DP approach is explained concisely. Notation is consistent and easy to follow: }''
 \noindent
  \underline{\emph{Logical Validity}}: ``\texttt{The algorithm is correct and would work, no logical flaw in the structure.}''
  
   \noindent
   \underline{\emph{Completeness}}: ``\texttt{the algorithm is complete and the answer is stated, but the computation is omitted. It's partial.}''

 \noindent
     \underline{\emph{Clarity}}: ``\texttt{The steps are well-explained and clear.}''
      
\vspace{5pt}

\noindent \textbf{Evaluation of the Baseline.} The baseline produces a low-quality proof, receiving a score of 0.19.

% `` \texttt{The proof contains significant logical errors, ... The transformation in Step 3 is flawed, ... fails to establish a true bijection. Additionally, the numerical inconsistency (Step 4 sums to 200 but Step 5 claims 2304) invalidates the conclusion. Thus, logical validity is low.} ''
  \noindent
\underline{\emph{Logical Validity}}: ``\texttt{The proof is logically invalid due to an incorrect reduction to only right and up moves, a flawed bijection in Step 3 and the contradiction between Step 4 and Step 5.}''

\noindent
     \underline{\emph{Completeness}}: ``\texttt{The proof is incomplete as it misses paths using diagonal moves, does not account for variable path lengths, and fails to properly handle the no-revisit constraint.}''

\noindent
\underline{\emph{Clarity}}: ``\texttt{The proof has a structured outline, but the marking system in Step 3 is ambiguous, and the numerical inconsistency creates confusion.}''
      
In summary, our method helps the model establish a solid logical ability and cultivate its insight to plan for the reasoning path.

\vspace{-0.2cm}

\section{Prompt Templates}
\label{prompts}
 We present all prompts on the remaining pages.

% \section{Statement on the Use of AI-Assisted Tools}
% We used AI-assisted tools to support the writing 
% and editing of this manuscript. Specifically, these 
% tools were employed to help refine grammar, improve readability, and enhance the clarity and consistency of the presentation. All technical content including the problem formulation, algorithm design, theoretical analysis, experimental setup, results, and conclusions, was developed and verified
% by the authors. The authors reviewed, validated,
% and edited all AI-assisted outputs to ensure correctness, originality, and compliance with the venue’s 
% policies.

\begin{figure*}
\begin{evaluationbox}{Insight Generation}
    Now you are tested the insight this math question: \\
    \{question\} \\
    From now on you can not go on thinking and expanding the proof step by step. Instead, try to infer the most potential core techniques that would be used in this question by ONLY analyzing the conditions in the question. You can only have a short and quick guess without detailed and long deduction. 
    
    \medskip
    The core techniques may refer to specific mathematical construction, use of known theorem or existing results or smart and subtle mathematical transformations, instead of those fundamental logic deduction. 
    
    \medskip
    Output: Latex formatted core precise techniques in your prediction for around 3 sentences.
\end{evaluationbox}
\end{figure*}

\begin{figure*}
\begin{evaluationbox}{Insightfulness Evaluation}
    The core techniques in math proof may refer to specific mathematical construction, use of known theorem or existed results or smart and subtle mathematical transformations for particular question, instead of those fundamental mathematical and logic deduction details. Now you have the following question: \\
    \{\}.
    
    \smallskip
    And a given insight which contains the prediction of core techniques to this question: \\
    \{\}
    
    \smallskip
    Now you need to review and evaluate this insight from the following aspects:
    \begin{enumerate}[label=\arabic*., leftmargin=1.5em, itemsep=0.2em, topsep=0.3em]
        \item see if there exists an idea in this insight that is highly key to the solution, which in general is likely to be found or realized after several thinking steps, or the other steps in the solution are far easier after dealing with this core idea. Note that such idea can not vaguely hints without providing concrete methods and identifying the core non-trivial step. And hence determine whether the insight is shallow quick guess or deep identification.
        
        \item see if the core ideas described in the insight is accurate enough. Check if the idea gives the precise techniques key and adapted particularly to the question. It does not need to be containing details but accurately describe mentioned techniques. And hence determine whether the idea is a simple scratch or an accurate expression.
        
        \item see if the core techniques mentioned in this insight are all core techniques for the question. check whether the question can be solved by filling basic mathematical and logical deductions ONLY under the core techniques in the insight. And hence determine whether the insight is comprehensive or incomplete.
    \end{enumerate}

    % 下半部分内容（对应第二张图）

    \vspace{0.8em}
    
    --output: \\
    For 1, if all ideas in the insight are not shallow quick guess, and there are no flaws in the whole insight, then output 'deep identification'. If there are also some ideas that are shallow quick guess, or even there are some flaws in this insight, output 'mixed'. Note that mixed means there indeed exist non-shallow ideas, but also other standard observations. Otherwise, if there are all shallow quick guess or just spread out standard general techniques in the insight without an accurate orientation, then output 'shallow quick guess'.
    
    \smallskip
    For 2, if all ideas in the insight are satisfied, then output 'detailed expression'. If all are not satisfied, output 'simple scratch'. otherwise output 'mixed'.
    
    \smallskip
    For 3, if all ideas in the insight are enough core ideas for the question, then output 'comprehensive'. If not, output 'incomplete'.
    
    \smallskip
    format: 
    \begin{enumerate}[label=\arabic*., leftmargin=4em, nosep]
        \item 'deep identification'/ 'shallow quick guess'/'mixed'
        \item 'detailed expression'/'simple scratch'/'mixed'
        \item 'comprehensive'/'incomplete'/'mixed'
    \end{enumerate}
\end{evaluationbox}
\end{figure*}

\begin{figure*}
\begin{evaluationbox}{Data Construction}
    Analyze the mathematical problem and solution below: \\
    \textbf{Problem:} \\
    \{question\} \\
    \textbf{Solution:} \\
    \{response\}
    
    \medskip
    Perform these tasks:
    
    1. First identify 1-3 core mathematical techniques used in the solution by considering if there are some specific constructions, theorem or existed results calling and smart mathematical transformations, where the smart transformation may not be known results and are subtle and hard to note.
    
    \smallskip
    Note that the core mathematical techniques are not those basic logic deductions. They need to be crucial to the solution.
    
    \smallskip
    Then Write the analysis for each technique on how a person can realize such technique when reading the question. The analysis should be from several aspects, including how the problem settings or assumptions suggest this technique, how it might be potentially useful to prove the result and why it might be crucial to the whole proof.
    
    \smallskip
    The analysis should be like an insightful and experienced math professor's thoughts when he is trying to solve the question. The whole analysis should start with 'Let's analyze the conditions in this question.'
    
    \smallskip
    The analysis contains 2-3 sentences for each technique. The analysis should avoid mentioning the solution. Make sure the logic of the analysis is coherent. After analysis, write the extracted techniques following the analysis through 'Therefore, the potential techniques are summarized as...', with three tags:
    
    \smallskip
    \textless construction\textgreater: identify the specific construction used in the solution if any; \\
    \textless theorem call\textgreater: specify the theorem or any existed results used in the solution if any; \\
    \textless transformation\textgreater: specify the smart mathematical transformations used in the solution if any.
    
    \smallskip
    If there are no such techniques, just specify 'no' after the tag, but you must write the tag \textless construction\textgreater, \textless theorem call\textgreater\ and \textless transformation\textgreater\ even there are no such techniques.
    
    \smallskip
    Do not write one technique under two tags both. The whole technique analysis, e.g., the analysis and technique extraction, should be with LaTeX inline math (\$... \textnormal{\$} ) where appropriate.
    
    \medskip
    2. Create a proof sketch integrating these techniques analyzed from task 1, which serves as a high-level proof organization:
    \begin{itemize}[label=-, leftmargin=2em, nosep]
        \item Format: Numbered steps in LaTeX
        \item Each step: 1 sentence with key mathematical reasoning
        \item Include essential formulas in math mode
        \item Example: '\textbackslash\textbackslash begin\{\{enumerate\}\}\textbackslash\textbackslash item Assume \$P\$ is countable: \$P = \textbackslash\textbackslash\{\{x\_1, x\_2, \textbackslash\textbackslash \dots \textbackslash\textbackslash\}\}\$\textbackslash\textbackslash end\{\{enumerate\}\}'
    \end{itemize}
    
    % \medskip
    % 3. Based on the original given solution, improve the solution by elaborating each step in the proof sketch, making it well-organized.
    
    % \medskip
    % 4. Output:
    % \begin{itemize}[label=-, leftmargin=2em, nosep]
    %     \item mathematical techniques: string (LaTeX formatted)
    %     \item proof sketch: string (LaTeX enumerated steps)
    %     \item solution: string (LaTeX formatted)
    % \end{itemize}
    
    % \smallskip
    % The mathematical techniques are enclosed within \textless tech\textgreater\textless /tech\textgreater, the proof sketch is enclosed within \textless sketch\textgreater\textless /sketch\textgreater\ and the solution is enclosed within \textless proof\textgreater\textless /proof\textgreater, respectively, i.e., \textless tech\textgreater\ mathematical technique analysis here \textless /tech\textgreater\ \textless sketch\textgreater\ proof sketch here \textless /sketch\textgreater\ \textless proof\textgreater\ solution here \textless /proof\textgreater
\end{evaluationbox}
\end{figure*}

\begin{figure*}

\begin{evaluationbox}{Data Construction (cont'd)}[t]
    3. Based on the original given solution, improve the solution by elaborating each step in the proof sketch, making it well-organized.
    
    \medskip
    4. Output:
    \begin{itemize}[label=-, leftmargin=2em, nosep]
        \item mathematical techniques: string (LaTeX formatted)
        \item proof sketch: string (LaTeX enumerated steps)
        \item solution: string (LaTeX formatted)
    \end{itemize}
    
    \smallskip
    The mathematical techniques are enclosed within \textless tech\textgreater\textless /tech\textgreater, the proof sketch is enclosed within \textless sketch\textgreater\textless /sketch\textgreater\ and the solution is enclosed within \textless proof\textgreater\textless /proof\textgreater, respectively, i.e., \textless tech\textgreater\ mathematical technique analysis here \textless /tech\textgreater\ \textless sketch\textgreater\ proof sketch here \textless /sketch\textgreater\ \textless proof\textgreater\ solution here \textless /proof\textgreater
\end{evaluationbox}

\end{figure*}

% 使用 figure* [t] 确保合并后的长盒子横跨双栏并置于页面顶部，绝不遮挡文字
\begin{figure*}
\begin{evaluationbox}{Proof Evaluation}
    The following question asks to prove a statement. \\
    The question: \\
    \{question\} \\
    The test subject's solution: \\
    \{response\}
    
    \medskip
    Your task is to evaluate the proof's quality and assign a score from 0 to 1 based on three criteria: logical validity (40\%), completeness (30\%), and clarity (30\%).
    
    \smallskip
    Instructions:
    \begin{enumerate}[label=\arabic*., leftmargin=1.5em, nosep]
        \item Analyze the proof step by step.
        \item For each criterion:
        \begin{itemize}[label=-, leftmargin=1.5em, nosep]
            \item Logical Validity: Check if each step follows logically from the previous one. Flag any logical errors.
            \item Completeness: Verify if all necessary cases and steps are included to prove the theorem.
            \item Clarity: Assess if the proof is clear, unambiguous, and well-explained.
        \end{itemize}
        \item Assign a sub-score (0 to 1) for each criterion and compute the total score using the weights: (0.4 $\times$ validity) + (0.3 $\times$ completeness) + (0.3 $\times$ clarity).
        \item Provide a brief explanation (2-3 sentences) summarizing any errors or issues and justifying the score.
    \end{enumerate}

    \vspace{0.8em}

    Output:
    \begin{itemize}[label=-, leftmargin=1.5em, nosep]
        \item Your total score: float;
        \item Your sub-scores and corresponding brief explanation:
        \begin{itemize}[label=-, leftmargin=1.5em, nosep]
            \item Sub-score: float;
            \item Brief explanation: string (LaTeX formatted)
        \end{itemize}
    \end{itemize}
    
    \smallskip
    The total score is enclosed in \textless score\textgreater\textless /score\textgreater, and the sub-scores with corresponding explanation are enclosed in \textless exp\textgreater\textless /exp\textgreater, e.g.,
    
    \smallskip
    - Format: \\
    '\textless score\textgreater \\
    your final scores here. Just write a single number here. \\
    \textless /score\textgreater \\
    \textless exp\textgreater \\
    "validity": your sub-score for validity here \\
    explanation: your explanation for validity score here \\
    "completeness": your sub-score for completeness here \\
    explanation: your explanation for completeness score here \\
    "clarity": your sub-score for clarity here \\
    explanation: your explanation for clarity score here \\
    \textless /exp\textgreater'
\end{evaluationbox}
\end{figure*}

\begin{figure*}
\begin{evaluationbox}{\texttt{InsightPO} Verifier Prompt}
    You are an expert mathematical proof evaluator specializing in insight-driven theorem proving. You will evaluate a model-generated response to a mathematical theorem-proving question.

    \medskip
    \textbf{Question:} \\
    \{question\} \\
    \textbf{Model's Response:} \\
    \{response\}

    \medskip
    \textbf{Evaluation Framework}

    \smallskip
    The response should follow a hierarchical insight-driven reasoning format containing three components:
    \begin{itemize}[label=-, leftmargin=1.5em, nosep]
        \item \textless tech\textgreater...\textless /tech\textgreater: Core technique analysis --- identifying key mathematical techniques by analyzing the problem's conditions
        \item \textless sketch\textgreater...\textless /sketch\textgreater: Proof sketch --- a high-level proof plan organized around the identified techniques
        \item \textless proof\textgreater...\textless /proof\textgreater: Full mathematical proof --- the complete, rigorous proof
    \end{itemize}

    \smallskip
    If the response does not contain this hierarchical structure, evaluate only the proof quality and assign 0 to Insight Quality.

    \medskip
    Evaluate the response across four dimensions:

    \smallskip
    \textbf{1. Insight Quality (Weight: 30\%)} \\
    Evaluate the core technique analysis in the \textless tech\textgreater\ section. Core techniques refer to specific mathematical constructions, invocations of known theorems or existing results, or clever mathematical transformations --- NOT basic logical deductions.

    \smallskip
    Check: Does the analysis begin by examining the conditions? Are the identified techniques specific, non-trivial, and genuinely crucial? Are techniques properly categorized under \textless construction\textgreater, \textless theorem call\textgreater, or \textless transformation\textgreater? Does the analysis demonstrate genuine mathematical intuition?

    \smallskip
    Scoring: 0.0--0.2: No meaningful techniques identified; 0.3--0.4: Only generic techniques; 0.5--0.6: Some correct techniques but missing key ones; 0.7--0.8: Most techniques correctly identified; 0.9--1.0: All essential techniques with deep analysis.

    \smallskip
    \textbf{2. Logical Validity (Weight: 30\%)} \\
    Evaluate mathematical rigor in the \textless proof\textgreater\ section. Check logical flow, justification of claims, absence of errors or circular reasoning.

    \smallskip
    Scoring: 0.0--0.2: Logically invalid; 0.3--0.4: Major logical errors; 0.5--0.6: Some gaps but sound direction; 0.7--0.8: Minor issues, fundamentally correct; 0.9--1.0: Every step rigorous.

    \smallskip
    \textbf{3. Completeness (Weight: 25\%)} \\
    Evaluate whether the proof fully establishes the theorem. Check coverage of all cases, edge conditions, and sufficiency of steps.

    \smallskip
    Scoring: 0.0--0.2: Absent or restates problem; 0.3--0.4: Significant portions missing; 0.5--0.6: Partial proof; 0.7--0.8: Nearly complete; 0.9--1.0: Fully complete.

\end{evaluationbox}
\end{figure*}

\begin{figure*}
\begin{evaluationbox}{\texttt{InsightPO} Verifier Prompt (cont'd)}
    \textbf{Scoring Instructions}

    \smallskip
    \textbf{4. Clarity (Weight: 15\%)} \\
    Evaluate presentation quality, notation consistency, and coherent flow of the hierarchical structure.
    
    \smallskip
    Scoring: 0.0--0.3: Incomprehensible; 0.4--0.6: Somewhat clear; 0.7--0.8: Clear with minor issues; 0.9--1.0: Exceptionally clear.
    
    \smallskip
    IMPORTANT: The final score MUST be exactly one of these 11 discrete values: \{0, 0.1, 0.2, 0.3, 0.4, 0.5, 0.6, 0.7, 0.8, 0.9, 1.0\}. No intermediate values are permitted.

    \smallskip
    \begin{enumerate}[label=\arabic*., leftmargin=1.5em, nosep]
        \item Analyze the response carefully, step by step.
        \item Assign a sub-score for each dimension from \{0, 0.1, 0.2, ..., 1.0\}.
        \item Compute the weighted total: raw $= 0.30 \times$ insight\_quality $+ 0.30 \times$ logical\_validity $+ 0.25 \times$ completeness $+ 0.15 \times$ clarity.
        \item Round the raw score to the nearest value in \{0, 0.1, ..., 1.0\} to obtain the final\_score.
    \end{enumerate}

    \medskip
    \textbf{Output Format}

    \smallskip
    '\textless score\textgreater \\
    \{final\_score --- must be one of: 0, 0.1, 0.2, 0.3, 0.4, 0.5, 0.6, 0.7, 0.8, 0.9, 1.0\} \\
    \textless /score\textgreater \\
    \textless exp\textgreater \\
    "insight\_quality": \{sub-score\} \\
    explanation: \{1-2 sentence justification\} \\
    "logical\_validity": \{sub-score\} \\
    explanation: \{1-2 sentence justification\} \\
    "completeness": \{sub-score\} \\
    explanation: \{1-2 sentence justification\} \\
    "clarity": \{sub-score\} \\
    explanation: \{1-2 sentence justification\} \\
    \textless /exp\textgreater'

\end{evaluationbox}
\end{figure*}

\begin{figure*}
\begin{evaluationbox}{Plan-and-Solve Baseline Prompt}
    You are given a mathematical theorem-proving problem. \\
    \textbf{Problem:} \\
    \{question\}

    \medskip
    First devise a concise proof plan. The plan should divide the theorem into necessary steps, identify useful intermediate claims, and specify how these claims will be connected. Do not assume access to any external proof sketch or annotated core techniques.

    \medskip
    Then solve the problem rigorously by following your plan. The final response should contain only the complete mathematical proof after the plan has guided your reasoning.

    \medskip
    Output the final proof enclosed by \textless proof\textgreater\textless /proof\textgreater.
\end{evaluationbox}
\end{figure*}

\begin{figure*}
\begin{evaluationbox}{Least-to-Most Prompt}
    You are given a mathematical theorem-proving problem. \\
    \textbf{Problem:} \\
    \{question\}

    \medskip
    Decompose the problem into a sequence of simpler subproblems or subclaims ordered from easiest to hardest. Each later subproblem may use the conclusions of previous subproblems. The decomposition must be generated only from the problem statement, without using any annotated proof sketch or core technique labels.

    \medskip
    Solve the subproblems one by one, and then assemble their conclusions into a complete proof of the original theorem. Make sure every intermediate claim used in the final proof is justified.

    \medskip
    Output the final proof enclosed by \textless proof\textgreater\textless /proof\textgreater.
\end{evaluationbox}
\end{figure*}

\begin{figure*}
\begin{evaluationbox}{SELF-DISCOVER Prompt}
    You are given a mathematical theorem-proving problem. \\
    \textbf{Problem:} \\
    \{question\}

    \medskip
    Select useful reasoning modules from the following general set:
    identify assumptions, infer hidden constraints, search for relevant lemmas, construct auxiliary objects, decompose into subclaims, prove intermediate claims, check edge cases, and assemble a final proof.

    \medskip
    Compose the selected modules into a task-specific reasoning structure for this problem. The structure should guide the proof process but must not rely on any annotated core techniques, proof sketches, or training-time hierarchy labels.

    \medskip
    Use the self-composed reasoning structure to generate a rigorous complete proof. Ensure that the final proof is logically valid, complete, and clearly written.

    \medskip
    Output the final proof enclosed by \textless proof\textgreater\textless /proof\textgreater.
\end{evaluationbox}
\end{figure*}

\end{document}